\documentclass{ieeeaccess}
\usepackage{cite}
\usepackage{amsmath,amssymb,amsfonts}
\usepackage{algorithmic}
\usepackage{graphicx}
\usepackage{textcomp}

\usepackage{bm}

\usepackage{booktabs}
\usepackage{multirow}
\usepackage[table]{xcolor}

\usepackage{graphicx}
\usepackage{amsmath}
\usepackage{amssymb}
\usepackage{booktabs}
\usepackage{multirow}
\usepackage{booktabs}
\usepackage[table]{xcolor}
\usepackage{comment}
\usepackage{stfloats} 
\usepackage{bm}
\usepackage{wrapfig}
\usepackage{subcaption}
\usepackage{subfig}

\makeatletter
\AtBeginDocument{\DeclareMathVersion{bold}
\SetSymbolFont{operators}{bold}{T1}{times}{b}{n}
\SetSymbolFont{NewLetters}{bold}{T1}{times}{b}{it}
\SetMathAlphabet{\mathrm}{bold}{T1}{times}{b}{n}
\SetMathAlphabet{\mathit}{bold}{T1}{times}{b}{it}
\SetMathAlphabet{\mathbf}{bold}{T1}{times}{b}{n}
\SetMathAlphabet{\mathtt}{bold}{OT1}{pcr}{b}{n}
\SetSymbolFont{symbols}{bold}{OMS}{cmsy}{b}{n}
\renewcommand\boldmath{\@nomath\boldmath\mathversion{bold}}}
\makeatother

\def\BibTeX{{\rm B\kern-.05em{\sc i\kern-.025em b}\kern-.08em
    T\kern-.1667em\lower.7ex\hbox{E}\kern-.125emX}}

\begin{document}
\history{Date of publication xxxx 00, 0000, date of current version xxxx 00, 0000.}
\doi{10.1109/ACCESS.2024.0429000}

\title{GoalNet: Goal Areas Oriented Pedestrian Trajectory Prediction}
\author{\uppercase{AMAR FADILLAH}\authorrefmark{1},
\uppercase{Ching-Lin Lee}\authorrefmark{2}, \uppercase{Zhi-Xuan Wang}\authorrefmark{3}, and \uppercase {Kuan-Ting Lai}\authorrefmark{4},
}

\address[1]{Department of Electronic Engineering, National Taipei University of Technology, Taipei 10608, Taiwan (t112999405@ntut.org.tw)}
\address[2]{Department of Electronic Engineering, National Taipei University of Technology, Taipei 10608, Taiwan (t111368001@ntut.org.tw)}
\address[3]{Department of Artificial Intelligence Technology, National Taipei University of Technology, Taipei 10608, Taiwan (t111c52026@ntut.org.tw)}
\address[4]{Department of Electronic Engineering, National Taipei University of Technology, Taipei 10608, Taiwan (ktlai@ntut.edu.tw)}

\markboth
{Author \headeretal: Preparation of Papers for IEEE TRANSACTIONS and JOURNALS}
{Author \headeretal: Preparation of Papers for IEEE TRANSACTIONS and JOURNALS}

\corresp{Corresponding author: Kuan-Ting Lai (e-mail: ktlai@ntut.edu.tw).}

\begin{abstract}
Predicting the future trajectories of pedestrians on the road is an important task for autonomous driving. The pedestrian trajectory prediction is affected by scene paths, pedestrian's intentions and decision-making, which is a multi-modal problem. \textcolor{black}{Relying solely on historical coordinates pedestrian data represents the most straightforward method for pedestrian trajectory prediction. Nevertheless, the accuracy achieved by this method is limited, primarily because it fails to account for the crucial scene paths impacting the pedestrian.} Instead of predicting the future trajectory directly, we propose to use scene context and observed trajectory to predict the goal points first, and then reuse the goal points to predict the future trajectories. By leveraging the information from scene context and observed trajectory, the uncertainty can be limited to a few target areas, which represent the "goals" of the pedestrians. In this paper, we propose GoalNet, a new trajectory prediction neural network based on the goal areas of a pedestrian. Our network can predict both pedestrian's trajectories and bounding boxes. The overall model is efficient and modular, and its outputs can be changed according to the usage scenario. Experimental results show that GoalNet significantly improves the previous state-of-the-art performance by 48.7\% on the JAAD and 40.8\% on the PIE dataset.
\end{abstract}

\begin{keywords}
pedestrians, trajectory prediction, future trajectories.
\end{keywords}

\titlepgskip=-21pt

\maketitle

\section{Introduction}
\label{sec:introduction}
\PARstart{T}rajectory prediction offers significant advantages across various domains by enabling proactive decision-making and enhancing safety. For example, in autonomous driving navigation, it is crucial for anticipating the movements of pedestrians and other vehicles, thereby facilitating collision avoidance and optimizing path planning. similarly, in vessel dynamic positioning \cite{kapal}, accurate trajectory forecasting supports precise station-keeping and safe maneuvering within complex maritime environments. Autonomous driving is under rapid technological development and changes in recent years, and the first priority must be safety. In order to achieve this goal, one of the fundamental tasks is to predict the trajectories of pedestrians, which can help autonomous vehicles to avoid collisions in advance. For example, if we cannot predict that a pedestrian is about to cross the road, the vehicle can only start braking when the pedestrian appears in front of it. The short response time may cause failure to stop in time, resulting in casualties. Therefore, estimating pedestrian trajectories quickly and accurately is a crucial task. But trajectory prediction is not an easy task, after all, people's actions and thoughts are constantly changing. There are two main factors that affect the choice of trajectory path. The first factor is the pedestrian's past trajectory, which contains the implicit intention. The second factor is the surrounding environment, including passable areas and inaccessible areas. A good pedestrian trajectory prediction model needs to consider these two factors.

\begin{figure}[tb]
    \centering
    \includegraphics[width=1\linewidth]{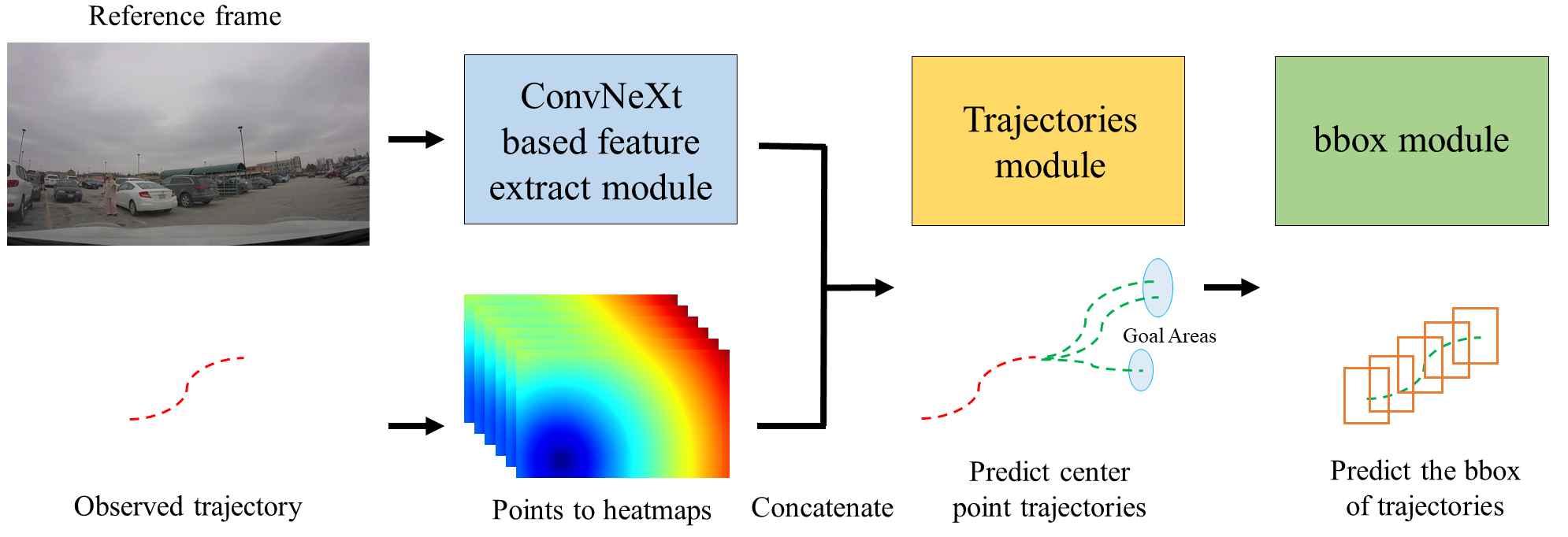}
    \caption{Overview of GoalNet. We use both scene context and observed trajectory to predict goal areas, and then use the goal information to predict pedestrian's trajectories and bounding boxes.}
    \label{fig:goalnet_arch}
\end{figure}

  Most of the previous trajectory prediction approach were based on the bird's eye view (BEV) datasets (e.g., ETH \cite{pellegrini2010improving}, UCY \cite{lerner2007crowds}, Standford Drone Dataset (SDD) \cite{Robicquet2016learning}). Because the pedestrian trajectory is predicted on the BEV, the 3-dimensional space task is transformed to 2-dimensional, and the fixed viewing angle indicates no self-motion, which simplifies the prediction problems in autonomous driving. In the autonomous driving system or the emergency braking system, the basis for judging the action command comes from the picture information from the perspective of the dashboard camera. Therefore, in the trajectory prediction of autonomous driving, the datasets from the perspective of the dashboard camera will be selected for research, such as JAAD \cite{rasouli2017JAAD} and PIE \cite{Rasouli2019PIE} datasets.

  \textcolor{black}{Trajectory prediction is frequently conceptualized as a time series analysis problem, which led early approaches to primarily rely on historical trajectories for input.}
  \textcolor{black}{However, for the purpose of trajectory prediction utilizing a full CNN architecture, it is essential to additionally consider the incorporation of scene context. This requires that the trajectory input and the two-dimensional environmental image be transformed to achieve a consistent dimensionality within the model's framework.}
  Since the scene image is 2-dimensional, and the trajectory is \textcolor{black}{1-dimensional}
  in the original data if the scene image is converted into the 1-dimensional alignment with the trajectory dimension, \textcolor{black}{the spatial information will be lost, but if the trajectory is converted into the 2-dimensional heat map to align with the scene image, this problem can be solved.}
  
  In terms of trajectory prediction output, deterministic trajectory prediction was mostly carried out in the early days, while multi-modal trajectory prediction is the mainstream of current trajectory prediction research\textcolor{black}{}.
  Because the pedestrian will change the route of the next action due to different environmental conditions or his ideas at each time step, predicting multiple uncertain trajectories can better represent pedestrian's future trajectory. Studies in \cite{anderson2019stochastic, yuning2020multipath, ivanovic2019trajectron, salzmann2020trajectron++} also show that multi-modal trajectory prediction is more accurate than deterministic trajectory prediction, and SGNet \cite{wang2022stepwise} has also achieved remarkable achievements in uncertainty prediction.

  Unlike the regression problem, which treats trajectories directly as sequential patterns, some previous studies, such as Y-Net \cite{mangalam2021goals}, proposed to address the problem in a novel way. The authors \textcolor{black}{argue}
  that pedestrian trajectory prediction is different from normal linear trajectory because the decision-making subject of trajectory is human, and it is obviously different from general time series forecasting because there are more potential variables of uncertainty in decision-making. The authors of Y-Net mentioned that such uncertain decision factors would have a greater impact on long-term trajectory prediction, and propose a long-term trajectory prediction network with good performance. Our work is inspired by Y-Net, but different from Y-Net's fixed BEV perspective. In our work, we make trajectory prediction based on vehicle \textcolor{black}{first person view (FPV)} perspective, which indicates that the decision variables of the trajectory besides the pedestrians, the driver's decision will have a similar impact on the future movement of the trajectory. And the speed of the vehicle will make these potential decision factors further amplified. Therefore, due to the more complex environmental conditions and the influence of more uncertain factors, the uncertainty trajectory prediction has become more important in our work.

  The contributions of this work are summarized as follows: First, we develop a new multi-modal trajectory prediction model GoalNet and show that it has a significant improvement in multi-modal trajectory prediction performance on JAAD and PIE trajectory prediction datasets. Second, different from the conditional variational autoencoder (CVAE) or RNN-based models in the field of trajectory prediction, we apply and integrate many advanced and new convolutional structures and methods. A well-performing design based on convolutional neural networks is proposed for trajectory prediction, and the influence of each sub-network in GoalNet has been studied. Finally, we provide a method to predict the corresponding bounding box trajectory from the center point trajectory. Figure \ref{fig:goalnet_arch} shows the overview structure of GoalNet.

\section{Related Work}
Nowadays, in the field of autonomous driving, the datasets used can mainly divided into two types. One is the datasets \textcolor{black}{containing LiDAR data},
such as nuScenes \cite{caesar2020nuscenes} or Argoverse\cite{chang2019argoverse}. After using LiDAR data, the dataset can be divided into two types. The ability to 3D detect and track the detected object and to measure its velocity and acceleration makes it possible to increase the accuracy of training autonomous driving-related neural networks by adding effective auxiliary information. 
\textcolor{black}{}Some recent studies such as ViP3D \cite{gu2023vip3d} have achieved remarkable results \textcolor{black}{on nuScenes dataset.} The other is a dataset with only camera images, such as JAAD and PIE, which are cheaper, smaller, and easier to install on vehicles than LiDAR. Especially in recent years, most vehicles have installed dashboard cameras. This means that the trajectory prediction model designed based on the scenarios of such datasets has a broader and universal application prospect. 
\textcolor{black}{When the neural network is designed to forego the processing of LiDAR data, it consequently reduces memory resource consumption and offers improved computational efficiency during execution.} For most vehicle safety assistance systems with limited hardware resources, these are the characteristics of such embedded systems with extremely limited computing resources.
  
  Early trajectory prediction was mostly studied in fixed-angle bird's eye view, but in BEV the 3D spatial task is simplified to 2D, and the fixed-angle means no self-motion, which simplifies the problem, however, to use it in autonomous driving, it is aided by the traffic recorder image, so recent research has also developed towards the first person view, and our GoalNet also uses FPV dataset to validate the model.
  
  \subsection{Bird's Eye View (BEV).} \textcolor{black}{Predicting pedestrian trajectories using a static single camera's Bird's-Eye View (BEV) primarily simplifies the challenge of accounting for self-motion. In this setup, trajectory prediction happens from a fixed viewpoint and position. Common datasets for this approach include ETH/UCY and SDD.}
  In ETH/UCY, the Y-Net \cite{mangalam2021goals} proposed by Mangalam et al. contains RGB scenes and historical trajectories as inputs, semantic segmentation of the scenes and estimation of trajectory endpoint locations and the probability distribution of the path points in them are used for trajectory prediction;The $V^{2}$-Net \cite{wong2022view} proposed by Wong et al. adds Fourier transformations to model and predict the trajectories in the spectral trajectory view, a kind of Transformer-based network containing coarse-level keypoints estimation and fine-level spectrum interpolation, this has led to an emphasis on pedestrian socialization and interaction with the scene; 
  \textcolor{black}{Goal-driven Long-Term Trajectory Prediction \cite{goal_add} proposed by Chen et al. introduces a method that enhances long-term forecasting by explicitly predicting a set of likely future goals for agents, subsequently generating diverse trajectories conditioned on these predicted goals. Its specialty lies in a dual-channel neural network architecture—comprising a Goal channel and a Trajectory channel—which employs Gated Recurrent Units (GRUs) to model temporal dependencies, automatically identifies and ranks destinations within the scene, and allows for generalization and transfer learning to unseen environments;}
  The NSP-SFM \cite{yue2022human} proposed by Yue et al. uses a deterministic model based on microscopic equations, and the key parameters can be learned from the data and integrated into the NSP model using an autoencoder that captures the kinematic dynamics and uncertainty. Trajectory prediction in a top-down view such as the Standford Drone Dataset has also yielded better results in the $V^{2}$-Net model and NSP-SFM mentioned above. In addition, the TDOR \cite{guo2022end} proposed by Guo et al. establishes the CVAE by taking the future trajectories in the neighboring grids as interpretable potential variables. They are iterated in maximum-entropy inverse reinforcement learning through differentiable value iteration networks and to maximize the data possibilities for learning path planning and trajectory generation. The PPNet  introduce a modular framework named PPNet for predicting pedestrian trajectories using multiple modalities. To improve prediction accuracy, PPNet incorporates a goal generator that models various distributions of potential goals, allowing it to identify and categorize diverse latent intentions. With the guidance of these goals, the anchor generator utilizes Transformer encoder-decoder networks to create anchors. Ultimately, the proposal generator combines the current position, anchors, and goals to produce a set of smooth, efficient, and probabilistic proposals. \cite{9989439} The combination of YOLO, DeepSORT, and LSTM effectively addresses the issues of occlusion and interactions between pedestrians, which are primary causes of errors in trajectory prediction. YOLO is used for accurate pedestrian detection, while DeepSORT helps with robust tracking by maintaining consistent identities even when pedestrians overlap or move in close proximity. Meanwhile, LSTM models integrated dot product attention mechanism with with the temporal dependencies in the trajectories, allowing the system to predict future positions more accurately despite these challenges. This integrated approach helps minimize errors caused by occlusions and interactions, improving overall trajectory prediction performance. \cite{Liu2023}

\begin{figure}[tb]
    \centering
    \includegraphics[width=1\linewidth]{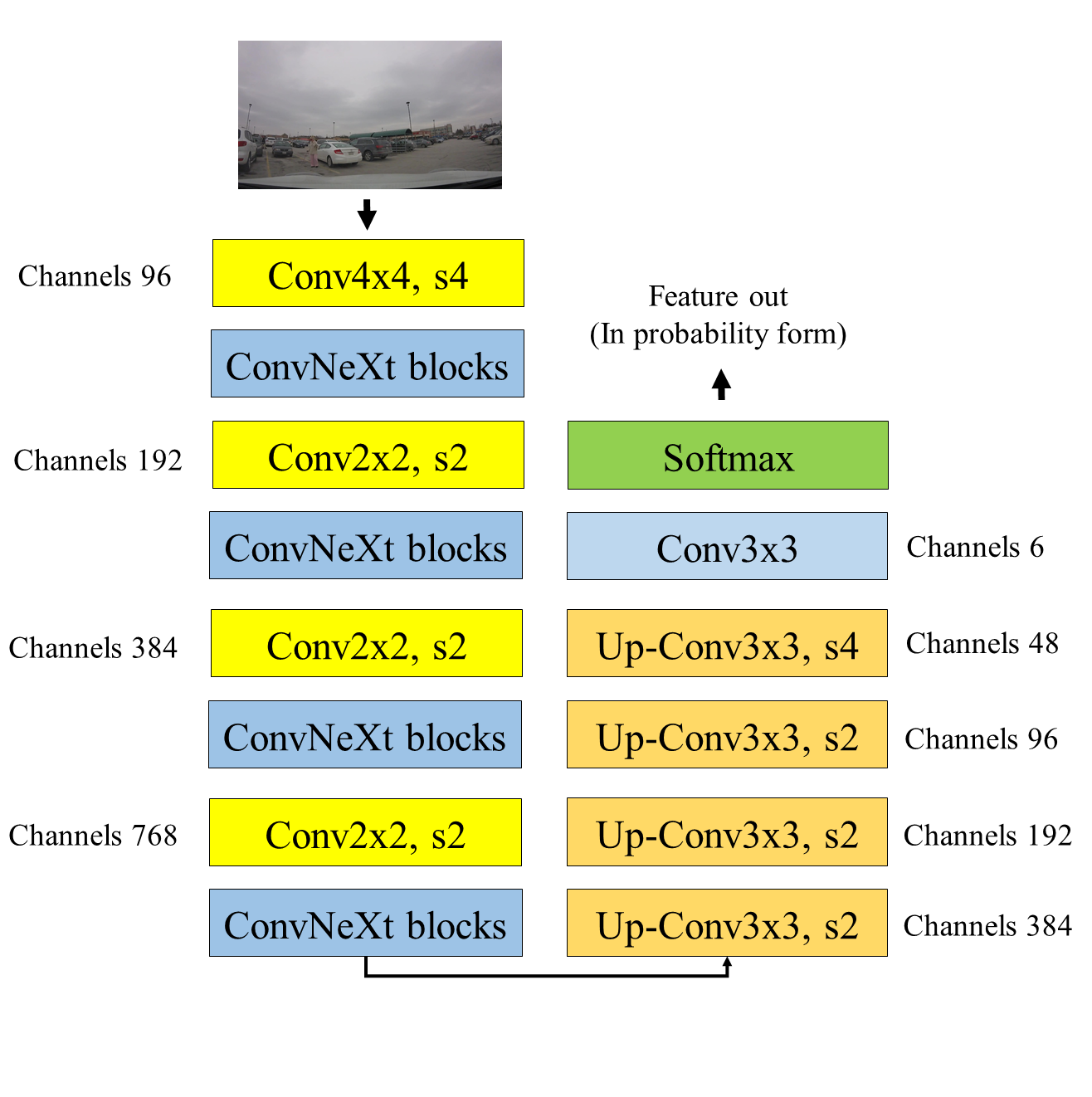}
    \caption{\textbf{GoalNet feature extract module architecture.} It consists of a ConvNeXt-T based encoder and a lightweight decoder based on Upsample convolution.}
    \label{fig:goalnet_feature}
\end{figure}

  \subsection{First Person View (FPV)} Pedestrian trajectory prediction with a first person perspective is more suitable for training automated driving systems, because it includes the vehicle's self-motion, in which JAAD and PIE datasets are more widely used, and we also used these two datasets for the validation. In 2019 the PIE dataset was released by Rasouli et al. They also proposed an intention estimation and trajectory prediction model with a large number of LSTM modules \cite{Rasouli2019PIE}, which achieved good results not only on its own dataset but also on JAAD. Yao et al. proposed the \textcolor{black}{BiTraP}
  model \cite{yao2021bitrap}, this trajectory prediction model consists of four parts: the condition prior network, the identification network, the target generation network and the trajectory generation network, and uses a large number of GRU modules in the model, It is a multi-modal trajectory prediction model that uses bounding boxes or trajectories as inputs and outputs multiple predicted trajectories. Recently Wang et al. indicated that human intention is a manifestation of planning, and intention at previous time points also changes their view in the present, and proposed SGNet \cite{wang2022stepwise}, which combines past goals into an encoder through average pooling, and uses an attention mechanism to learn the importance of each goal to the prediction. And both Bitrap and SGNet perform well on JAAD and PIE dataset. Several methods emphasize the pose representation of pedestrians. Zhou et al. integrated Bayesian Networks (BN) and SWI-PROLOG with the knowledge graph technique to infer three types of intentions that influence pedestrian crossing trajectory prediction. Subsequently, pedestrian intentions, historical trajectories, and body posture angle features were combined for trajectory prediction using a Bi-LSTM+Attention network. The enhanced Casrel model, along with pedestrian crossing intention estimation and trajectory prediction, was validated through experimental evaluation.The experiments demonstrated that the Casrel model successfully extracted triplet information, while the BN intention inference model effectively predicted pedestrian crossing intentions. \cite{Zhou2023}

\begin{figure*}[tb]
    \centering
    \includegraphics[width=1\linewidth]{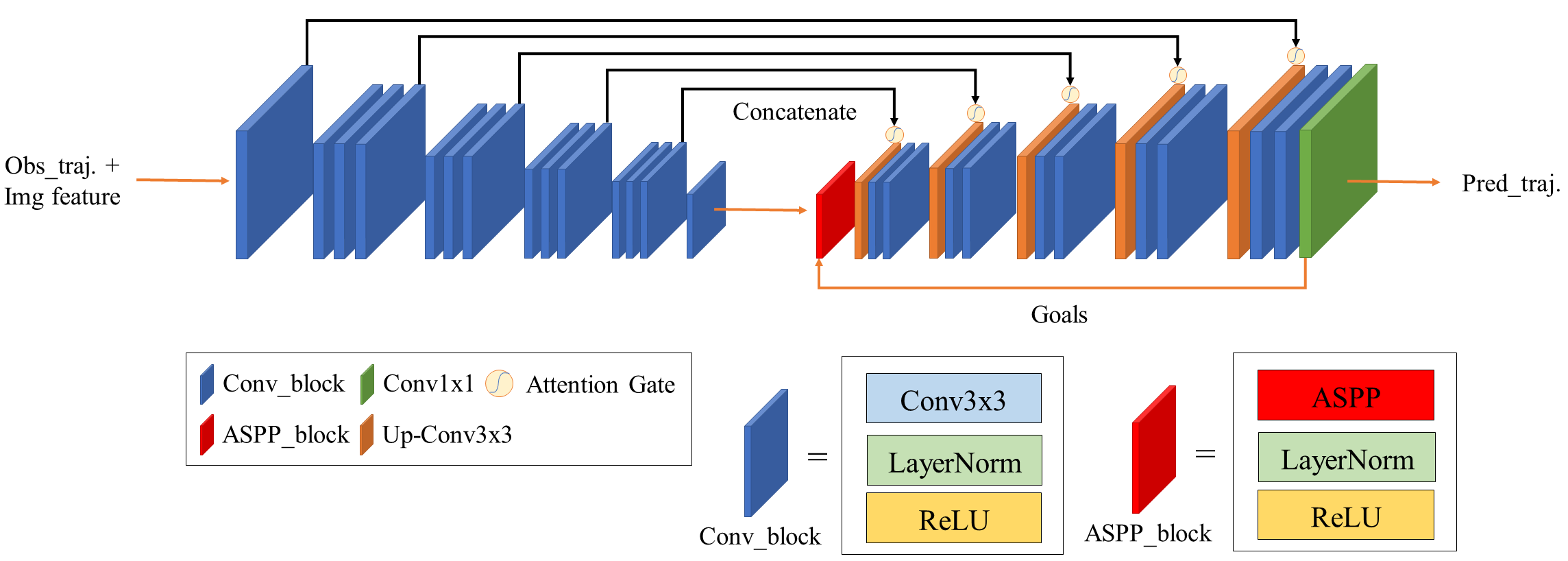}
    \caption{Trajectory module designs for GoalNet.}
    \label{fig:goalnet_traj}
\end{figure*}
\section{Proposed Methods}
The problem of trajectory prediction can be defined as follows. Suppose that the current time is \emph{t}, and \textbf{P} represents the bounding box coordinates of the object with length and width (in pixels), then \textbf{P}{\small{\emph{t}}} represents the position of the object at time \emph{t}. The given observation time is 15 frames \textcolor{black}{or 0.5 seconds in 30 frames per seconds (FPS) video, \textbf{X} = \{\textbf{P}{\small{\emph{t}-14}}, \textbf{P}{\small{\emph{t}-13}}, ..., \textbf{P}{\small{\emph{t}}}\}, our goal is to predict the position of \textbf{X} in the next 45 frames or 1.5 seconds, \textbf{Y} = \{\textbf{P}{\small{\emph{t}+1}}, \textbf{P}{\small{\emph{t}+2}}, ..., \textbf{P}{\small{\emph{t}+45}}\}.} For the prediction of stochastic trajectories, given an observation trajectory \textbf{X}, there may be multiple results, so various trajectories corresponding to these uncertain futures need to be generated to cover this uncertainty, the \textcolor{black}{multi-modal trajectories }
are expressed as \textbf{S}, and the corresponding \emph{n} trajectories are generated. \textbf{S} = \{\textbf{S}{\small{1}}, \textbf{S}{\small{2}}, ..., \textbf{S}{\small{\emph{n}}}\}.
  
  GoalNet consists of three modules, which are feature extraction, trajectory prediction, and bounding box prediction module, and we estimate the number of parameters and FLOPs of our model as shown in Table \ref{tab_params}.

\begin{table}[htb]
\centering
    \begin{tabular}{lcr}
        modules & \#param. & FLOPs\\
        \toprule
        GoalNet - extract & 31.347M & 22.770G \\
        GoalNet - trajectory & 2.131M & 20.821G \\
        GoalNet - bbox & 0.174M & 0.174M \\
        \midrule
        Total & 33.652M & 43.591G \\
        \bottomrule
    \end{tabular}
\caption{GoalNet \#param. \& FLOPs.}
\label{tab_params}
\end{table}

\subsection{Feature Extract Module}
\textcolor{black}{Before scene context was widely used as important data for trajectory prediction, some previous studies \cite{Alahi2016Social, yao2021bitrap, wang2022stepwise, gupta2018social} dealt with trajectories without incorporating environmental images.} 
  On specific trajectory datasets (e.g., Standford Drone Dataset (SDD) \cite{Robicquet2016learning}), some studies \cite{liang2020garden, liang2020simaug} begin to use segmentation as an environmental feature extraction method. However, for most trajectory prediction datasets, In particular, it is not an easy task to construct segmentation annotation for first person perspective (FPV) datasets, \textcolor{black}{}
  it is difficult to define appropriate classes, and the output channels are not very flexible, because they depend on the number of segmentation classes. If RGB scene images are directly input into the trajectory prediction model, the trajectory prediction model needs to adapt to the processing of environmental images and trajectory information at the same time. In order to make the trajectory prediction model consider both scene context and trajectory information, the trajectory prediction module focuses on trajectory coding. Rather than directly input RGB images into the trajectory prediction model, it is necessary to establish the image and maintain the hidden state of complete spatial information through feature extraction. Indirectly inputting the scene image into the trajectory prediction module is a better method.

  We implemented the feature extraction module based on ConvNeXt \cite{liu2022convnet}, the structure of which is shown in Figure \ref{fig:goalnet_feature}. Considering the computational load of this module and the trajectory prediction module, we chose the smallest ConvNeXt-Tiny as the encoder. After extracting the feature map through ConvNeXt-Tiny, the decoder uses the same but greatly simplified steps as downsample to upsample back the feature map to the size of the original RGB image and normalizes the features to probability distribution through softmax, because softmax can enlarge the feature differences. Highlight important feature areas within each channel.

\begin{figure*}[tb]
    \centering
    \includegraphics[width=1\linewidth]{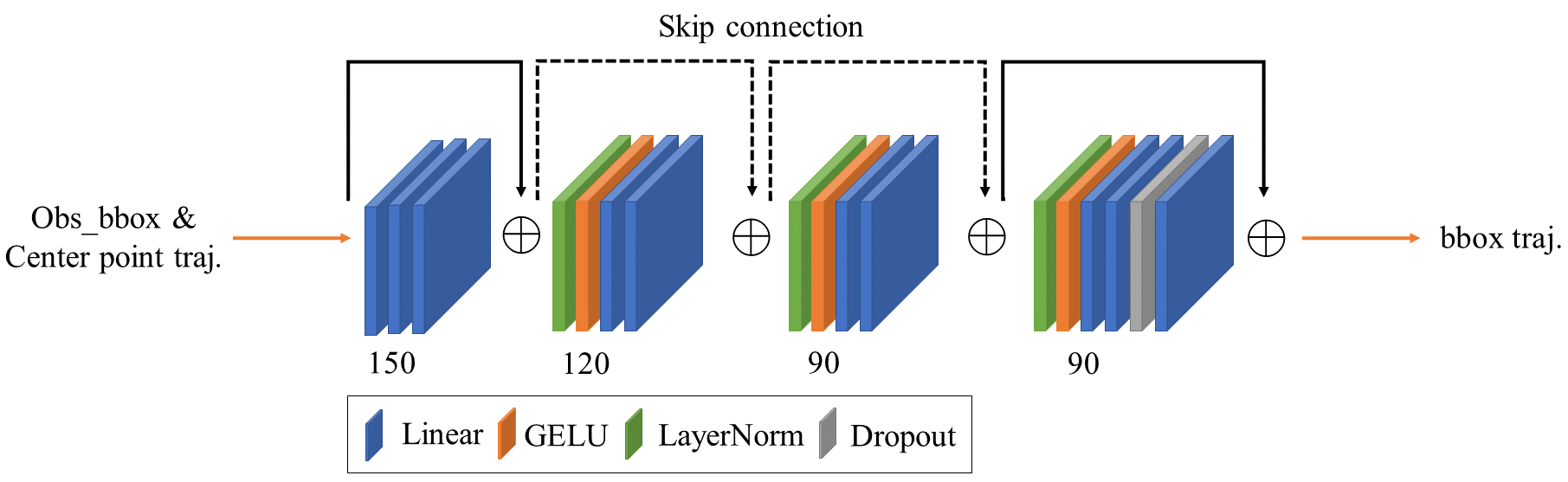}
    \caption{\textbf{bbox module architecture of GoalNet.} Solid line indicates that the skip connection is a direct connection. Dashed lines indicate that the number of channels in the skip connection has been converted.}
    \label{fig:goalnet_bbox}
\end{figure*}

\begin{figure}[tb]
    \centering
    \includegraphics[width=1\linewidth]{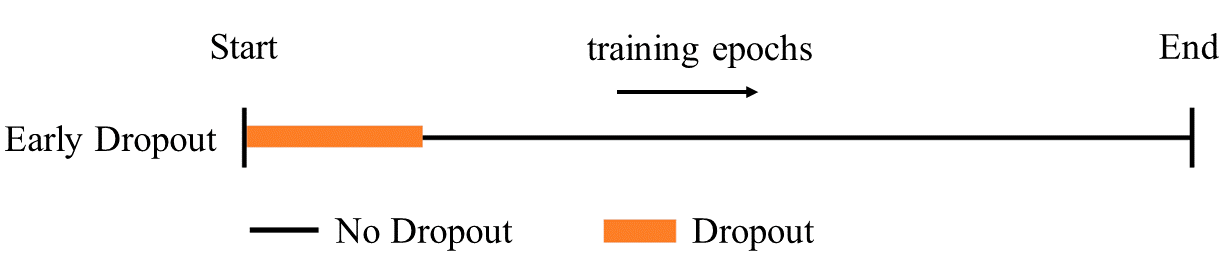}
    \caption{\textbf{Early Dropout.} This method improves the generalization and final accuracy of the model.}
    \label{fig:early_dropout}
\end{figure}

\subsection{Trajectory Module}
  In the trajectory prediction module, the same two-dimensional processing method as the environmental image can avoid the loss of spatial information contained in the environmental image. Some previous studies \cite{tang2022evostgat, xue2020poppl, czech2022onboard} mostly adopted the RNN-based trajectory prediction model. However, when the environmental image is encoded as a one-dimensional hidden state, the spatial information contained in the environmental image will be completely destroyed. Therefore, we converted the trajectory to the heatmap form to transform the trajectory into a two-dimensional space representation aligned with the representation of the scene image.

  This module encodes the observed trajectory and the scene image, and the decoder predicts the goal areas represented by a probability distribution, and calculates the main goal through spatial soft arg-max \cite{levine2016end}. At the same time, K-means clustering algorithm is used to calculate the secondary goals in the goal areas, and these goals are input back to the decoder in sequence again to generate \textcolor{black}{multi-modal}
  trajectories with these goals as future assumptions.
  
  As a classic convolutional encoder-decoder network structure, U-Net \cite{ronneberger2015u} was originally designed for segmentation. Still, the trajectory is highly correlated with space, so designing a structure similar to U-Net is potentially beneficial. Y-Net \cite{mangalam2021goals} has proven this.
  
  Batch Normalization (BN) \cite{ioffe2017batch} was used in the U-Net, but BN of trajectories will corrupt the information. So BN should be replaced with a batch independent normalization method, like Group normalization (GN) \cite{wu2018group} or Layer Normalization (LN) \cite{ba2016layer}, The GN restriction that channels must be divisible by the group severely limits model design, the combination of LN and Atrous Spatial Pyramid Pooling (ASPP) \cite{chen2017deeplab} matches better. So we chose to use LN to substitute the BN.
  
  Since trajectory prediction does not require a special ability to perceive the edge of an object like segmentation, stride 2 convolution, like modern convolutional networks such as ResNet \cite{he2016deep} and ConvNeXt \cite{liu2022convnet} is probably more advantageous in this case. Therefore, we replace the U-Net downsample method with stride 2 convolution. At the same time, by replacing the convolution layer with ASPP at the deepest (Center layer) of the trajectory prediction module, \textcolor{black}{receptive field}
  of the model is greatly improved, and the spatial information of the trajectory is more fully utilized.
  
  In addition, the Attention Gate (AG) technique is proposed in attention U-Net \cite{oktay2018attention}, and it has been proved that AG can enhance attention to important regions in feature maps concatenated by the encoder. Since our trajectory prediction module is similar to U-Net and AG is not designed for segmentation, but belongs to a more general technology, we add AG to our trajectory prediction module. The final trajectory prediction module structure is shown in Figure \ref{fig:goalnet_traj}.

\subsection{Bounding Box Module}
  In Figure \ref{fig:goalnet_bbox}, in order to predict the bounding box corresponding to the trajectory, we convert the center point trajectory predicted by the trajectory prediction module to the trajectory in the form of bounding box by predicting the width and height of the bounding box. Different from previous studies \cite{yao2021bitrap, wang2022stepwise}, previous studies directly predicted bounding box location, coordinates sequences representing the upper left and lower right points of bounding box, and then calculated the trajectory of its center point. In our work, it is not easy to generate two points at a time in a single time step, so we adopt the opposite method, that is, first predict the location, coordinates of the center point, and then predict the trajectory in the form of bounding box. As shown in the table \ref{tab_params}, the computational consumption of this method can be almost ignored.
  
  This module uses the fully connected layer design, and PreAct ResNet \cite{he2016identity} proves that it is more beneficial to reduce the layers on skip connection that hinder the propagation of gradient flow, so the residual structure in our module is designed in a similar way to PreAct. In addition, recent studies have shown that the final accuracy of the model can be improved by using Early-Dropout \cite{liu2023dropout}, so we adopted this method (Figure \ref{fig:early_dropout}).

\subsection{Loss Functions}
\textcolor{black}{The final layer of the trajectory module generates a heatmap of probability distribution, which is localized by spatial soft arg-max. Consequently, binary cross-entropy loss is applied to supervise the module.}
  For the bounding box prediction module, since the prediction results are absolute values of width and height, we use 
  \textcolor{black}{SmoothL1 loss }
  \cite{girshick2015fast} to monitor. Thus, for each training sample, our final loss is summarized as follows,
\begin{eqnarray}
&\mathcal{L}_{traj} = BCEWithLogits(P(\bm{Y}), P(\bm{Y}_{gt})),\\
&\mathcal{L}_{bbox} = SmoothL1(\bm{Y}_{wh}, \bm{Y}_{wh\_gt}),\\
&\mathcal{L}_{total} = \mathcal{L}_{traj} + \mathcal{L}_{bbox},
\end{eqnarray}
  Where, P(\textbf{Y}) is the probability distribution of the predicted trajectory at each timestep in the future, and P({\textbf{Y}\small{\emph{gt}}}) is the heatmap transformed by the ground truth coordinate of the target object in the future. {\textbf{Y}\small{\emph{wh}}} is the width and height of the predicted trajectory bounding box at each point in time in the future, and {\textbf{Y}\small{\emph{wh\_gt}}} is the true size of the target object in the future.

\section{Experiments}
\label{sec:sxperiments}

\begin{table*}[tb]
\centering
    \begin{tabular}{lcccccccc}
        \toprule
        \multirow{3}*{Method (Best of 20)} & \multicolumn{4}{c}{JAAD} & \multicolumn{4}{c}{PIE} \\
        \cmidrule(lr){2-5}\cmidrule(lr){6-9}
        ~ & MSE↓ & C{\scriptsize{MSE}}↓ & CF{\scriptsize{MSE}}↓ & \multirow{2}*{NLL↓} & MSE↓ & C{\scriptsize{MSE}}↓ & CF{\scriptsize{MSE}}↓ & \multirow{2}*{NLL↓}\\
        ~ & (0.5s / 1.0s / 1.5s) & (1.5s) & (1.5s) & & (0.5s / 1.0s / 1.5s) & (1.5s) & (1.5s) \\
        \midrule
        BiTraP-GMM \cite{yao2021bitrap} & 153 / 250 / 585 & 501 & 998 & \textbf{16.0} & 38 / 90 / 209 & 171 & 368 & \textbf{13.8}\\
        BiTraP-NP \cite{yao2021bitrap} & 38 / 94 / 222 & 177 & 565 & 18.9 & 23 / 48 / 102 & 81 & 261 & 16.5\\
        SGNet-ED \cite{wang2022stepwise} & 37 / 86 / 197 & 146 & 443 & - & 16 / 39 / 88 & 66 & 206 & -\\
        ABC+ \cite{abc2022} & 40 / 89 / 189 & 145 & 409 & - & 16 / 38 / 87 & 65 & 191 & -\\
        \midrule
        GoalNet & \textbf{31} / \textbf{67} / \textbf{133} & \textbf{97} & \textbf{215} & 17.8 & \textbf{15} / \textbf{32} / \textbf{65} & \textbf{44} & \textbf{97} & 15.7\\
        \bottomrule
    \end{tabular}
\caption{Stochastic results (\emph{K}=20) on JAAD and PIE in terms of MSE/C{\scriptsize{MSE}}/CF{\scriptsize{MSE}}. ↓ denotes lower is better.}
\label{tab_result}
\end{table*}

\begin{table*}[tb]\scriptsize
\setlength{\belowcaptionskip}{0.2cm}
\centering
    \begin{subtable}[c]{0.48\textwidth}
        \centering
        \begin{tabular}{lccc}
        \multirow{2}*{Case} & MSE↓ & C\tiny{MSE}↓ & CF\tiny{MSE}↓ \\
        ~ & (0.5s / 1.0s / 1.5s) & (1.5s) & (1.5s) \\
        \hline
        w/o feature extract module & 34 / 72 / 144 & 108 & 272 \\
        w/o Attention-Gate & 33 / 72 / 139 & 103 & 229 \\
        w/o ASPP & 30 / 69 / 137 & 100 & 235 \\
        \rowcolor{gray!30}
        \cellcolor{white}GoalNet & \textbf{31} / \textbf{67} / \textbf{133} & \textbf{97} & \textbf{215} \\
        \end{tabular}
        \caption{Macro design.}
        \label{tab_explore:a}
    \end{subtable}
    \hfill
    \begin{subtable}[c]{0.48\textwidth}
        \centering
        \begin{tabular}{lccc}
        \multirow{2}*{Case} & MSE↓ & C\tiny{MSE}↓ & CF\tiny{MSE}↓ \\
        ~ & (0.5s / 1.0s / 1.5s) & (1.5s) & (1.5s) \\
        \hline
        w/o norm. & 30 / 67 / 136 & 100 & 249 \\
        BN & 114 / 235 / 446 & 403 & 1028 \\
        GN & 32 / 70 / 141 & 104 & 232 \\
        \rowcolor{gray!30}
        \cellcolor{white}LN & \textbf{31} / \textbf{67} / \textbf{133} & \textbf{97} & \textbf{215} \\
        \end{tabular}
        \caption{Feature normalization. LayerNorm outperforms other normalizations.}
        \label{tab_explore:b}
    \end{subtable}
    \begin{subtable}[c]{0.48\textwidth}
        \centering
        \begin{tabular}{lccc}
        \multirow{2}*{Case} & MSE↓ & C\tiny{MSE}↓ & CF\tiny{MSE}↓ \\
        ~ & (0.5s / 1.0s / 1.5s) & (1.5s) & (1.5s) \\
        \hline
        \rowcolor{gray!30}
        \cellcolor{white}ReLU & \textbf{31} / \textbf{67} / \textbf{133} & \textbf{97} & \textbf{215} \\
        Leaky ReLU(0.1) & 31 / 69 / 137 & 101 & 216 \\
        PReLU & 30 / 65 / 134 & 98 & 233 \\
        SiLU & 33 / 72 / 141 & 104 & 222 \\
        GELU & 32 / 72 / 149 & 111 & 267 \\
        \end{tabular}
        \caption{Activation function. ReLU is simple and efficient.}
        \label{tab_explore:c}
    \end{subtable}
    \hfill
    \begin{subtable}[c]{0.48\textwidth}
        \centering
        \begin{tabular}{lccc}
        \multirow{2}*{Case} & MSE↓ & C\tiny{MSE}↓ & CF\tiny{MSE}↓ \\
        ~ & (0.5s / 1.0s / 1.5s) & (1.5s) & (1.5s) \\
        \hline
        \rowcolor{gray!30}
        \cellcolor{white}Conv (stride 2) & \textbf{31} / \textbf{67} / \textbf{133} & \textbf{97} & \textbf{215} \\
        MaxPool & 30 / 67 / 134 & 97 & 221 \\
        AvgPool & 30 / 66 / 136 & 98 & 240 \\
         &  &  & \\
         &  &  & \\
        \end{tabular}
        \caption{Feature downsample.}
        \label{tab_explore:d}
    \end{subtable}
    \caption{Exploration study of our model on JAAD. ↓ denotes lower is better. Our final proposal is marked in \colorbox{gray!30}{gray}.}
    \label{tab_explore}
\end{table*}

  In this section, we use a total of two publicly available trajectory prediction datasets on pedestrian behavior in traffic to study the performance of GoalNet.
  
  \paragraph{Datasets.} Joint Attention in Autonomous Driving (JAAD) \cite{rasouli2017JAAD} and Pedestrian Intention Estimation (PIE) \cite{Rasouli2019PIE} Dataset are used in our experiment. Trajectories for both datasets were recorded using an on-board camera, recorded and annotated at 30 frames per second (FPS). JAAD contained 2,800 pedestrian trajectories out of 75,000 annotated frames, while PIE contained 1,800 pedestrian trajectories out of 293,000 annotated frames, with longer annotated trajectories and more comprehensive annotations than JAAD.
  We followed the JAAD and PIE standard training/testing split \cite{Rasouli2019PIE}, using the same observation and prediction lengths as prior work \cite{yao2021bitrap, wang2022stepwise}. A trajectory with a length of 0.5 seconds (15 frames) is input to generate a trajectory with a length of 0.5, 1.0, 1.5 seconds (15, 30, 45 frames).
  
  \paragraph{Implementation Details.} We use ConvNeXt-Tiny \cite{liu2022convnet} as the backbone of the feature extraction module, convert the two-dimensional input of RGB 3 channels into two-dimensional output of 6 channels, and implement the encoder-decoder of the trajectory prediction module using a U-Net-like structure. The input of the trajectory prediction module is taken by the observed 15-time step trajectory heatmap as 15 channels and concatenate with 6 channels scene feature map. The channel size of the encoder is 32-32-64-64-64-128, while the decoder is the opposite, and the trajectory prediction output of 45 time steps is generated. The last point of the first output trajectory is feedback to the decoder as a goal to generate the final future trajectories.
  \textcolor{black}{All models were trained on a single RTX2080Ti GPU using a batch size of 8, the Adam \cite{kingma2014adam} optimizer with an initial learning rate of 1e-4, and the ReduceLROnPlateau scheduler.}

  \paragraph{Evaluation Metrics.} Following \cite{rasouli2017JAAD, Rasouli2019PIE, yao2021bitrap, wang2022stepwise}, our GoalNet model uses Mean Square Error (MSE), Center Mean Square Error (CMSE), Center Final Mean Square Error (CFMSE) and Kernel Density Estimation-based Negative Log Likelihood (KDE-NLL) to evaluate our performance on JAAD and PIE in pixels. MSE is calculated according to the upper left and lower right coordinates of the bounding box trajectory. CMSE and CFMSE are calculated using the center point of the bounding box trajectory. Calculate the NLL of the ground truth under a probability distribution fitted by KDE using 2000 trajectory samples. \\ 
  \begin{figure*}[b]
    \centering
    \begin{subfigure}[b]{0.48\textwidth}
        \centering
        \includegraphics[width=\textwidth]{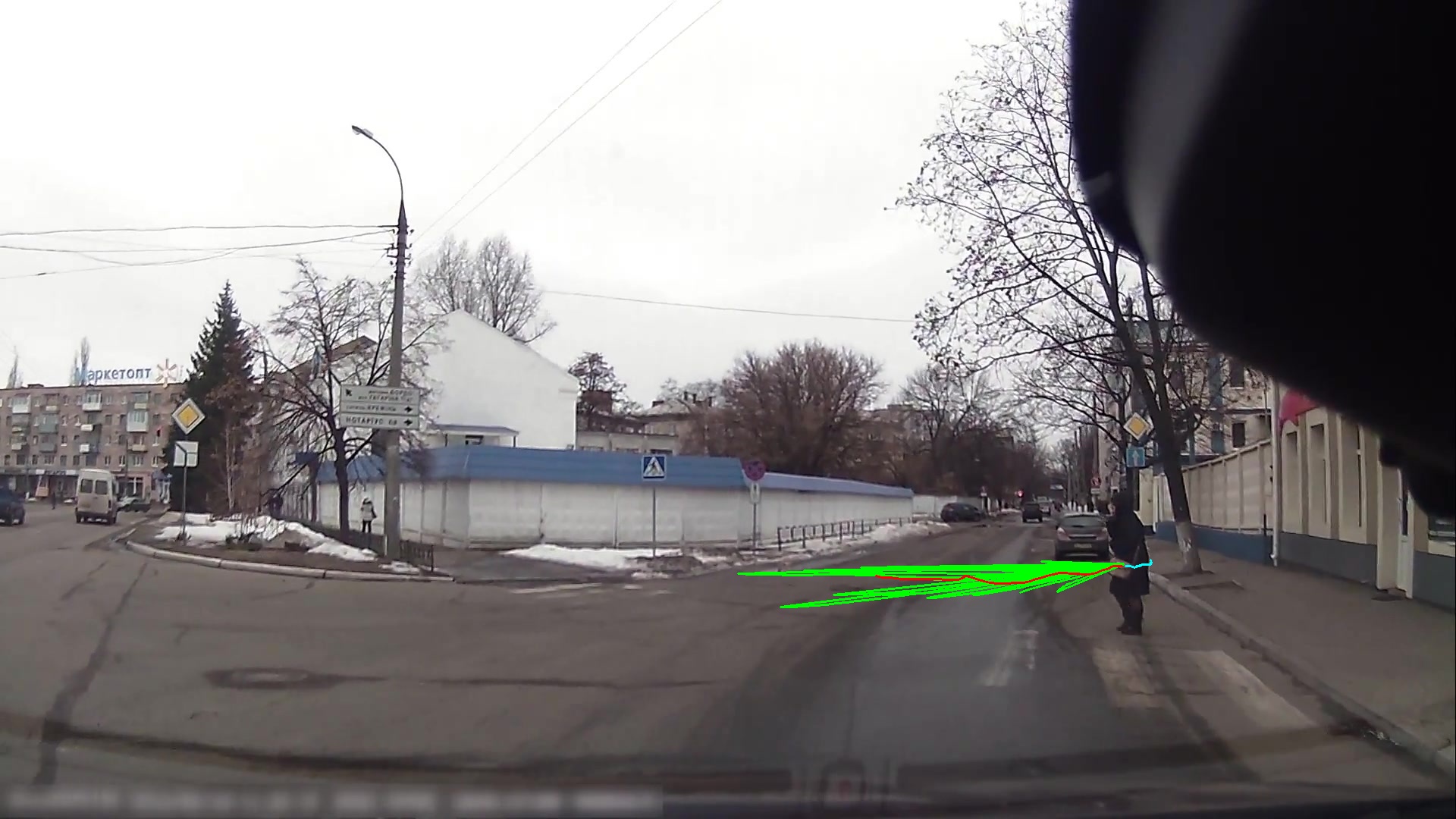}
    \end{subfigure}
    \hfill
    \begin{subfigure}[b]{0.48\textwidth}
        \centering
        \includegraphics[width=\textwidth]{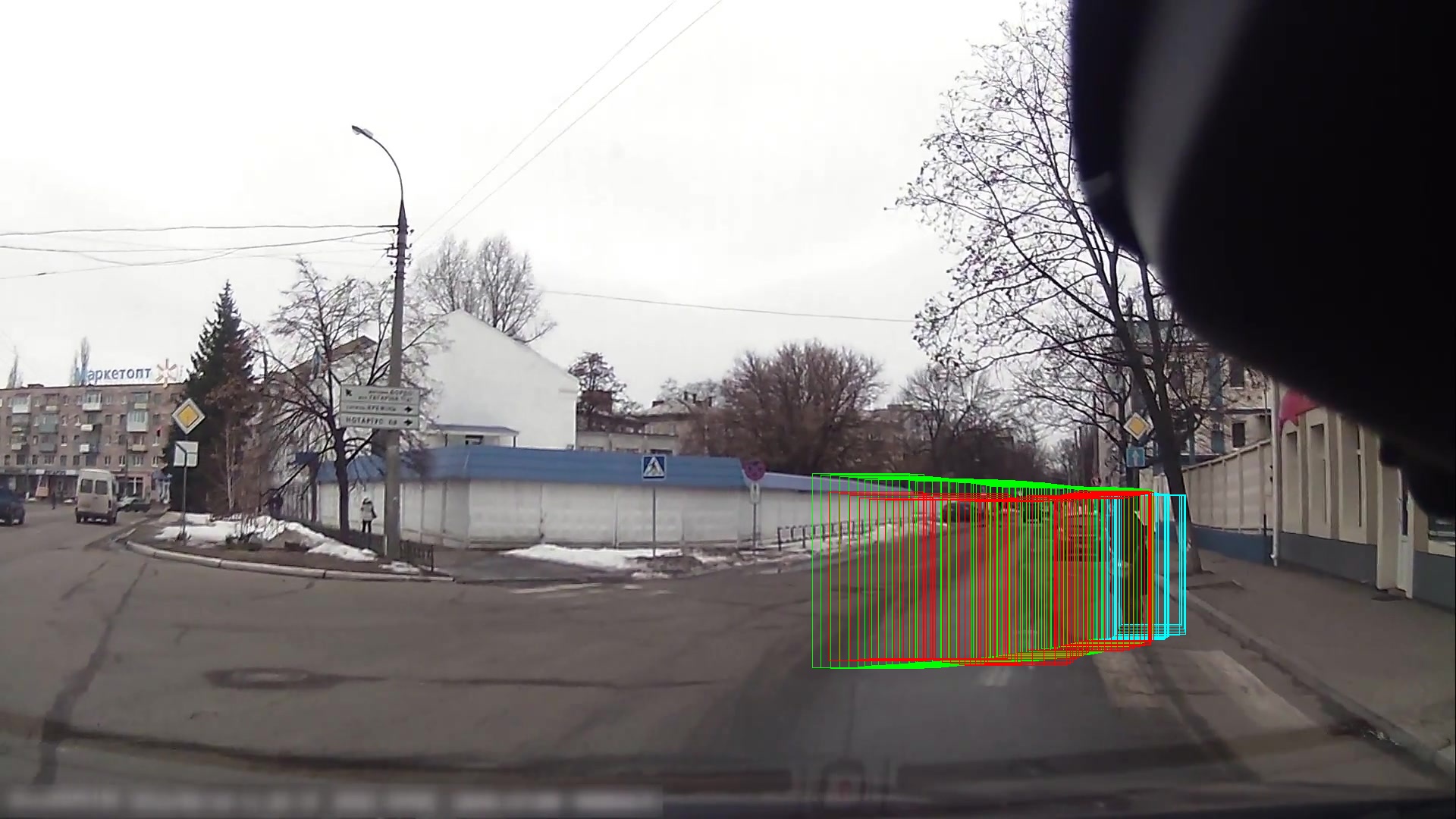}
    \end{subfigure}
    \\
    \begin{subfigure}[b]{0.48\textwidth}
        \centering
        \includegraphics[width=\textwidth]{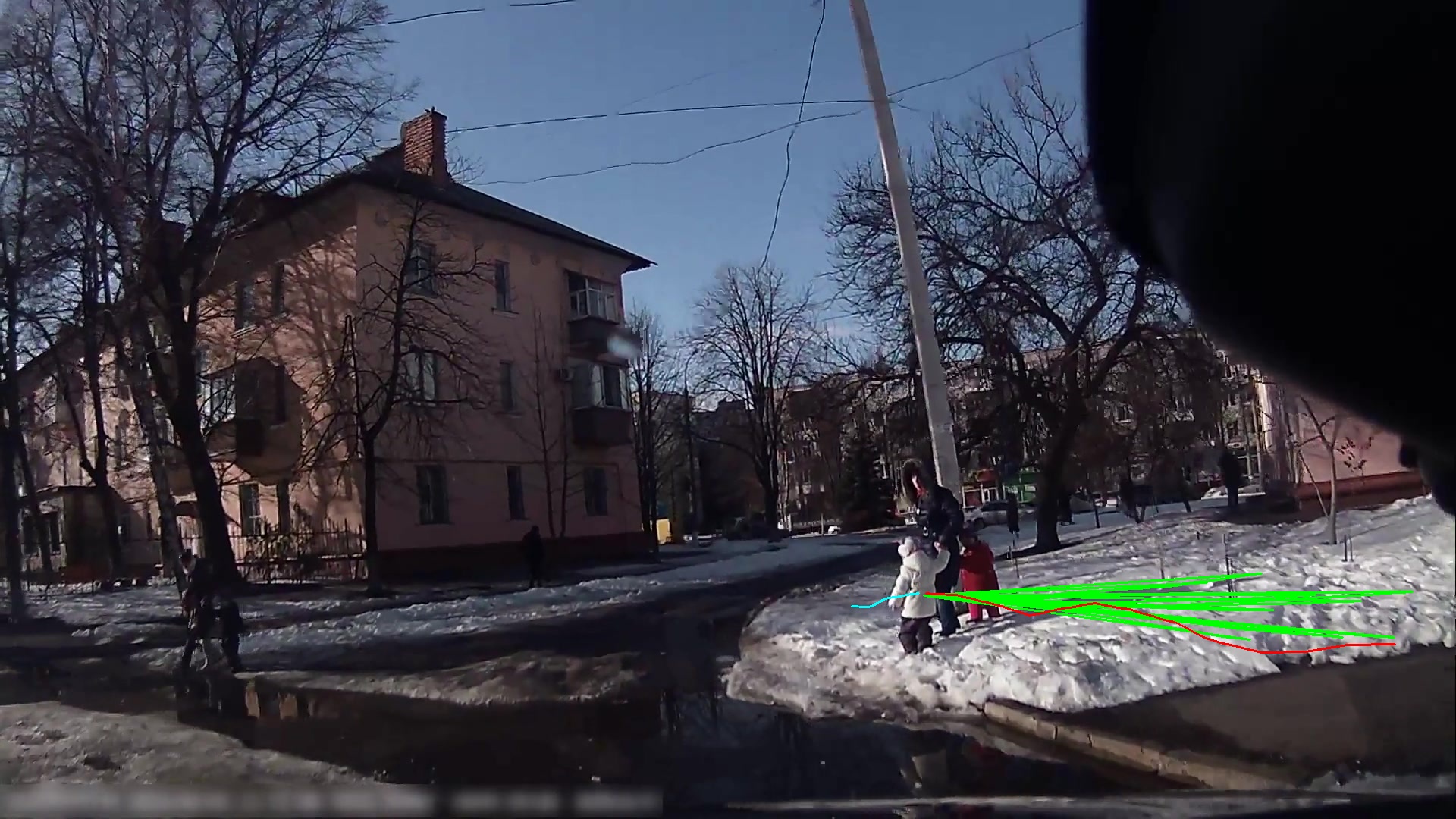}
        \caption{Center point trajectories.}
    \end{subfigure}
    \hfill
    \begin{subfigure}[b]{0.48\textwidth}
        \centering
        \includegraphics[width=\textwidth]{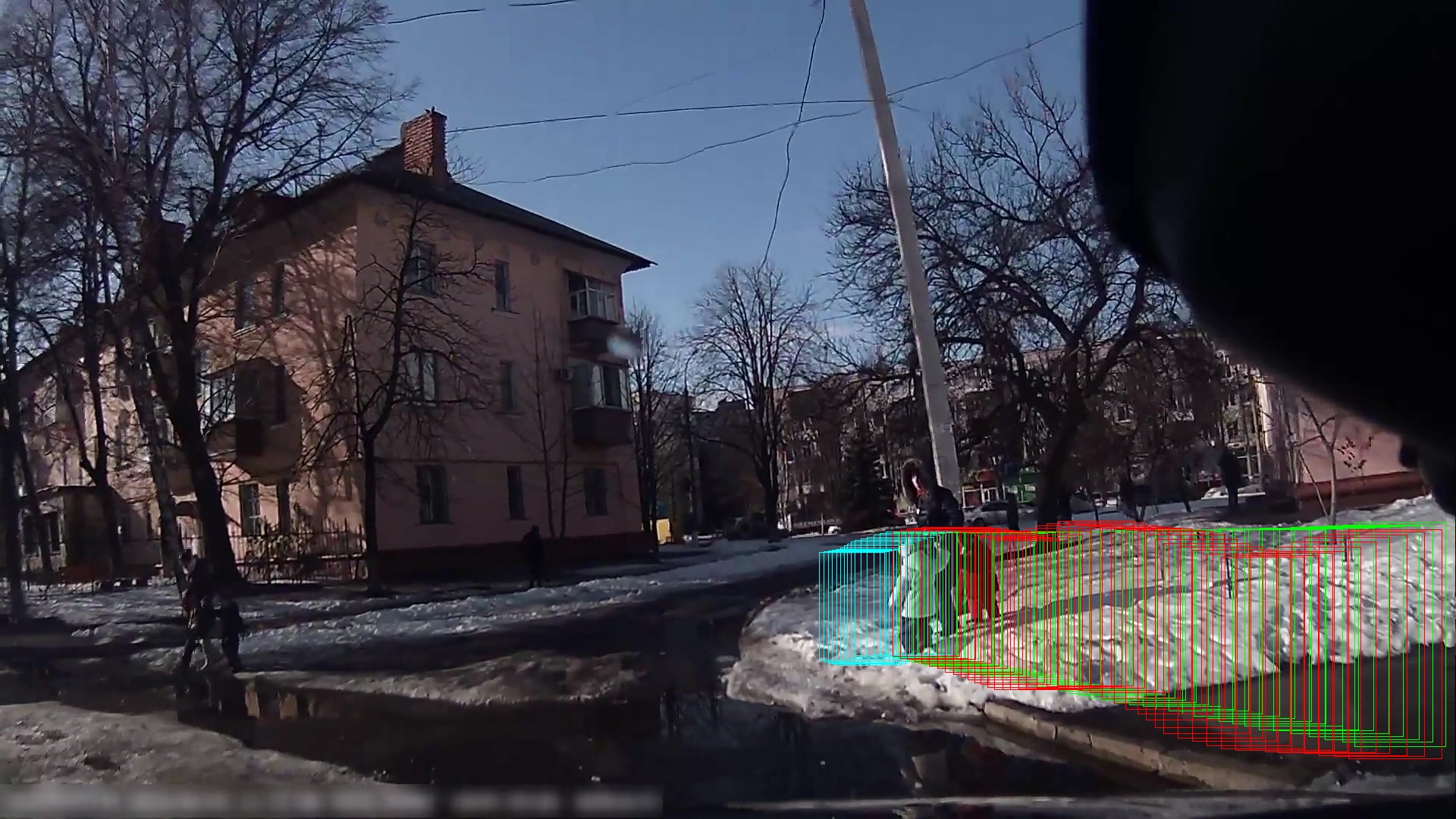}
        \caption{bbox trajectory.}
    \end{subfigure}
    \caption{\textbf{JAAD Qualitative Results.} The cyan color indicates the observed trajectory, the red color indicates the ground truth future trajectory, and the green color indicates the predictions from our GoalNet model.}
    \label{fig:jaad_viz}
\end{figure*}
\begin{figure*}[tb]
    \centering
    \begin{subfigure}[b]{0.48\textwidth}
        \centering
        \includegraphics[width=\textwidth]{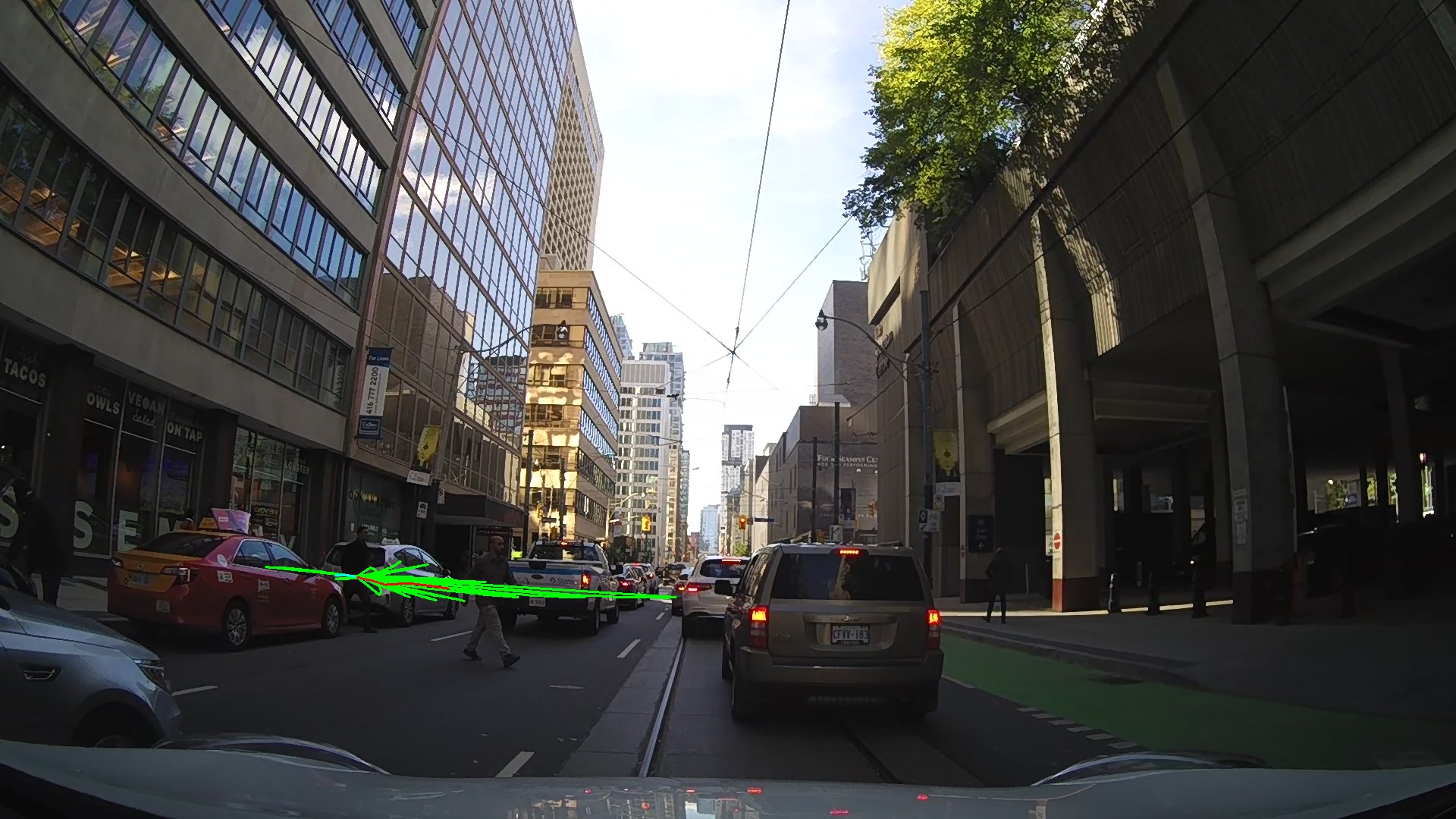}
    \end{subfigure}
    \hfill
    \begin{subfigure}[b]{0.48\textwidth}
        \centering
        \includegraphics[width=\textwidth]{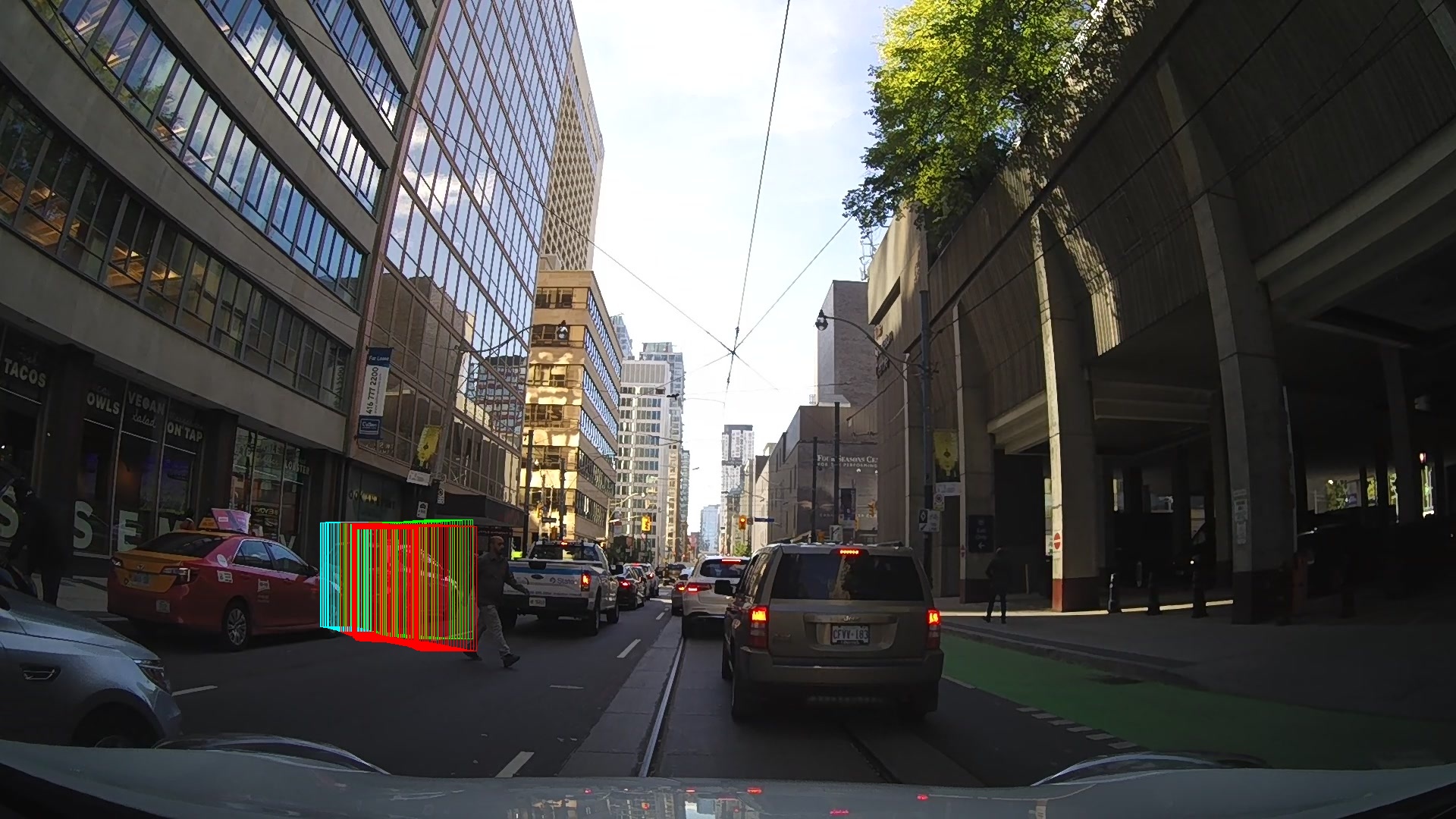}
    \end{subfigure}
    \\
    \begin{subfigure}[b]{0.48\textwidth}
        \centering
        \includegraphics[width=\textwidth]{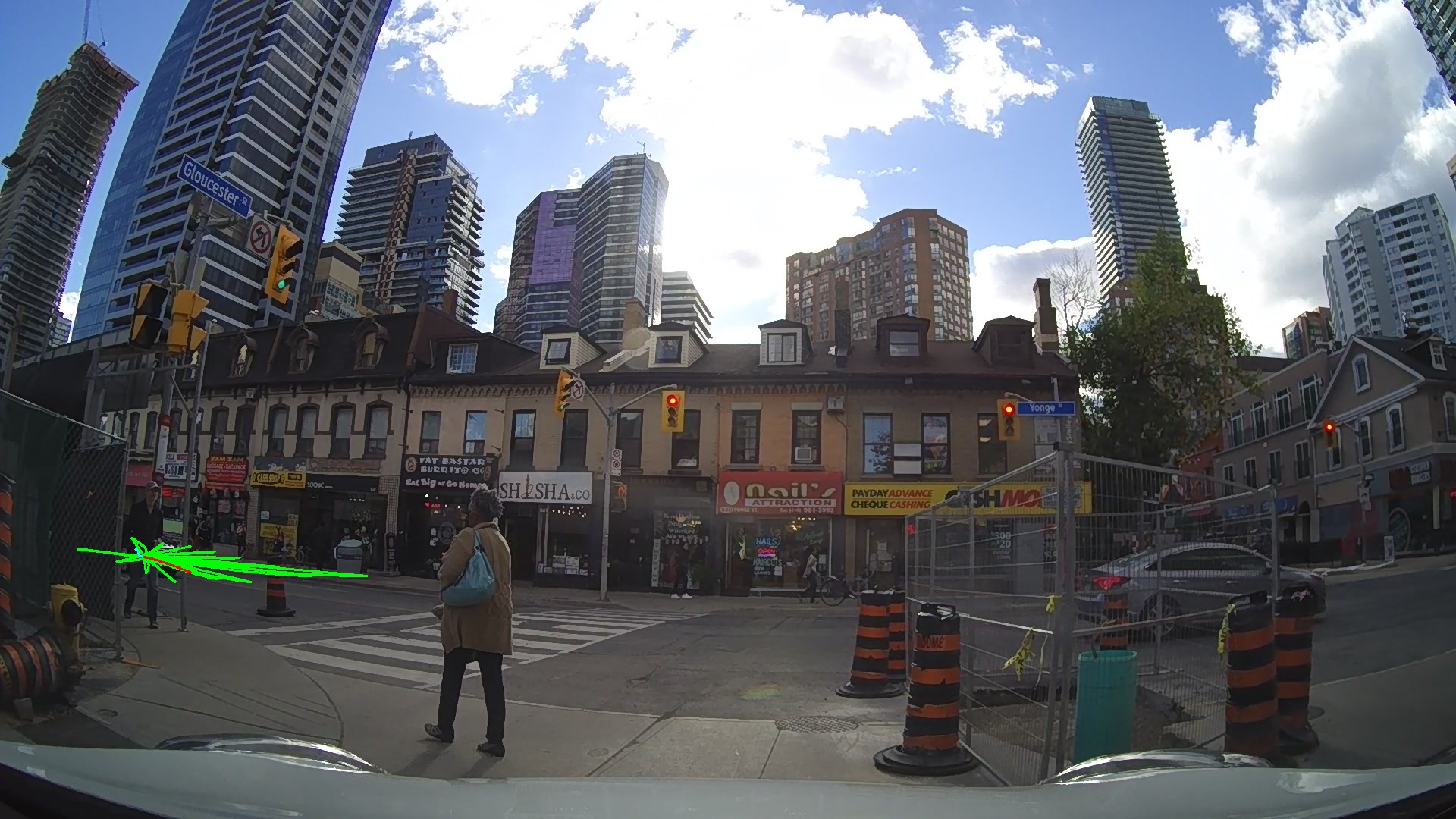}
        \caption{Center point trajectories.}
    \end{subfigure}
    \hfill
    \begin{subfigure}[b]{0.48\textwidth}
        \centering
        \includegraphics[width=\textwidth]{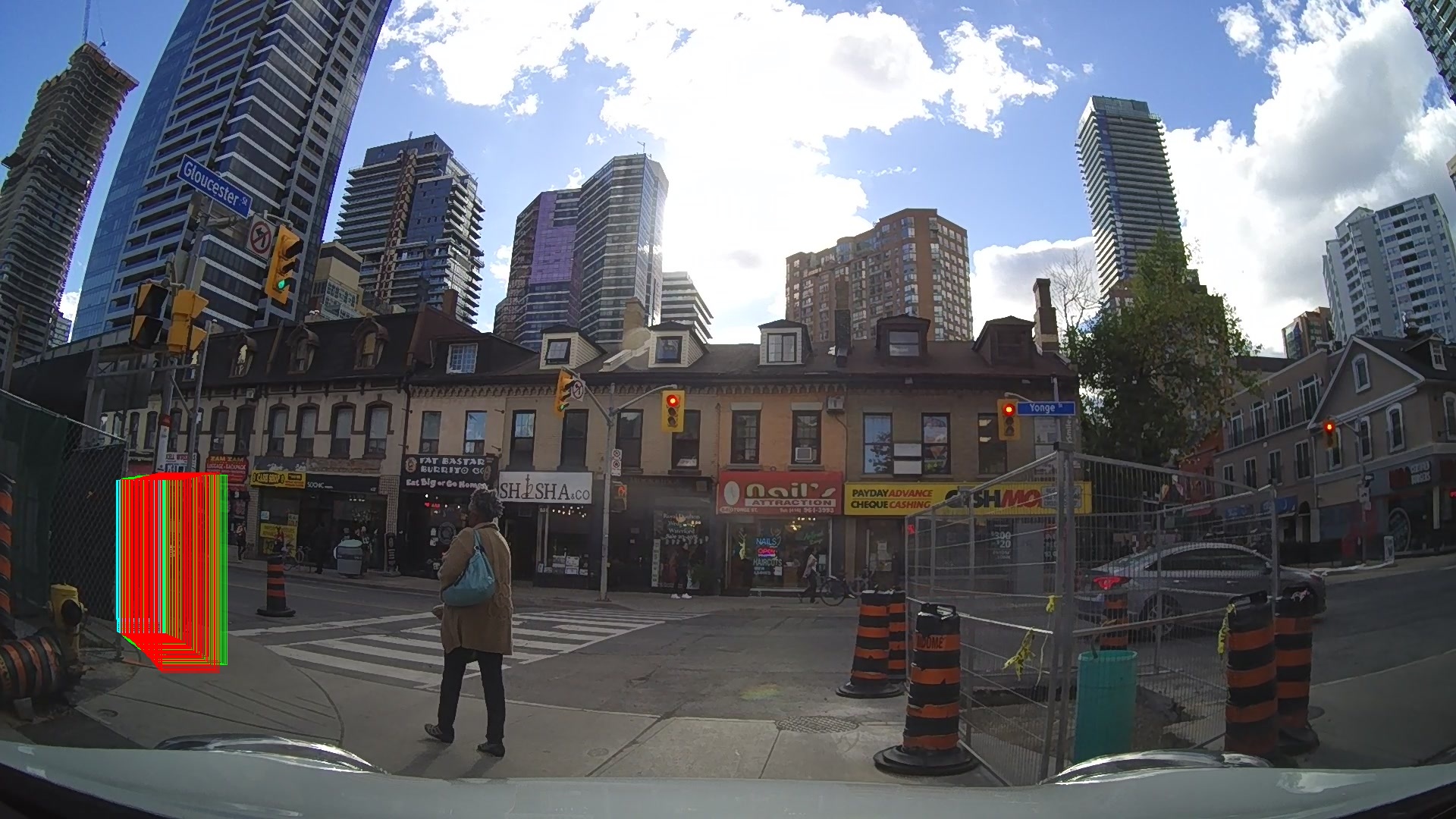}
        \caption{bbox trajectory.}
    \end{subfigure}
    \caption{\textbf{PIE Qualitative Results.} The cyan color indicates the observed trajectory, the red color indicates the ground truth future trajectory, and the green color indicates the predictions from our GoalNet model.}
    \label{fig:pie_viz}
\end{figure*}
\subsection{Experiments on JAAD and PIE datasets}
  We compared the results with the following baseline models: 1) BiTraP-GMM 2) BiTraP-NP 3) SGNet-ED. \textcolor{black}{To ensure a fair comparison with the stochastic results detailed in \cite{yao2021bitrap} and \cite{wang2022stepwise} for the JAAD and PIE Benchmarks, we select and present the most optimal trajectory from a collection of K=20 multi-modal generative trajectories.}
  Table \ref{tab_result} shows the trajectory prediction results of JAAD and PIE datasets. The results show that our method significantly outperforms previous state-of-the-art methods by an average of 48.7\% on JAAD and 40.8\% on PIE.

  Although the NLL metrics is not the best, the reason is that the multi-modal trajectory distribution generated by BiTraP-GMM is similar to ground truth, but each trajectory has a high displacement error. Our method has lower displacement error and better multi-modal trajectory distribution than BiTraP-NP.

\subsection{Exploration Study}

  To design our GoalNet model and make predictions on the JAAD and PIE datasets, we use the U-Net \cite{ronneberger2015u} architecture as a base, and reference modern convolutional neural network designs such as ConvNeXt \cite{liu2022convnet}. Evaluate the submodules and micro-designs of the overall GoalNet design, the result is shown in Table \ref{tab_explore}.
  
  Table \ref{tab_explore:a} analyzes the impact of removing different modules on GoalNet. The results show that the feature extraction module is necessary. If the environment RGB image is directly input into the trajectory prediction module, the final accuracy will be greatly reduced. And removing either the Attention Gate (AG) \cite{oktay2018attention} or Atrous Spatial Pyramid Pooling (ASPP) \cite{chen2017deeplab} modules would result in lower final accuracy, so adding AG and ASPP does have a positive effect on the model.

  Table \ref{tab_explore:b} analyzes the performance using different normalization layers, and while Batch Normalization (BN) \cite{ioffe2017batch} is an important and common component in convolutional neural networks and is preferred for many visual tasks, because it improves the convergence of the model and reduces overfitting, but experiments show that BN is very unsuitable for use in our structure, and the experimental data show that Group Normalization (GN) \cite{wu2018group} is better than BN. However, the implementation of layer normalization of GN is very similar to BN, the difference is that GN avoids some of the batch restrictions of BN by group normalization of channels within a single training sample, and we speculate that normalization on batches will destroy some important information in the training data. In recent years, Layer Normalization (LN) \cite{ba2016layer} has gradually increased in importance and replaced BN gradually, including replacing BN with LN in transformer and ConvNeXt. ConvNeXt shows that it is not difficult to replace BN with LN in convolutional neural networks, and it is possible to achieve better results. Our experiments also show that LN does perform best, so LN is used as the layer normalization choice in our model.

  The results of different activation functions are discussed in Table \ref{tab_explore:c}. Rectified Linear Unit (ReLU) is widely used in convolutional neural networks, and due to the dying ReLU problem, ReLU has therefore derived many variants (e.g., LeakyReLU, PReLU, GELU) to solve this problem. Experiments have shown that ReLU performs best in our model, so we have chosen to continue using ReLU in our model.
  
  The influence of different downsampling layers is studied in Table \ref{tab_explore:d}. Max Pooling is often used in previous networks, it retains key features by taking the maximum value during downsampling. In addition, Max Pooling's gradient flow propagation is easier than stride 2 convolution, which makes it easier for neural networks to learn. However, in modern convolutional neural networks, stride 2 convolution is often used to replace Pooling, because stride 2 convolution has learnable parameters. With appropriate design, some accuracy advantages may be achieved.

\subsection{Qualitative Results}
  Figure \ref{fig:jaad_viz} and Figure \ref{fig:pie_viz} show GoalNet's prediction results on the test split of the JAAD and PIE datasets, respectively. (a) Show the prediction results for all 20 \textcolor{black}{multi-modal}
  trajectories. Most of the multi-modal prediction results are close to the ground truth trajectory in the future, and the high probability distribution near the ground truth trajectory is successfully predicted, which also indicates the stability of the model and converges the uncertainty to a certain range. (b) The best bounding box trajectory is shown. The results show that our model can correctly predict the scaling of the bounding box as the pedestrian approaches or moves away from the vehicle.
  
  The predicted future trajectory in the scene not only shows the trajectory and intention of the pedestrian, but also includes the movement of the vehicle itself. The predicted result also represents the relative position of the pedestrian and vehicle after the specified time step. For the vehicle assistance system, the relative position of the pedestrian and the vehicle in the future is obviously helpful to the safety decision of the system.

\section{Conclusion}
We propose a multi-modal uncertainty trajectory prediction model called GoalNet based on convolutional neural networks. It is also demonstrated that our proposed model can achieve state-of-the-art results on first-person datasets suitable for autonomous driving training. Overall, GoalNet improved previous state-of-art performance on JAAD by 48.7\% and improved it by 40.8\% on PIE. For future work, we expect to further expand our model with optional modular modules such as social collisions, intentions, LiDAR, etc., so that the model can flexibly make predictions based on the information it has to further ensure universality while making the best use of available information to improve safety.

\section{Acknowledgment}
This research was supported by the National Science and Technology Council, Taiwan, under Grant No. NSTC 113-2221-E-027-048-MY3.


\newpage
\begin{IEEEbiography}[{\includegraphics[width=1in,height=1.25in,clip,keepaspectratio]{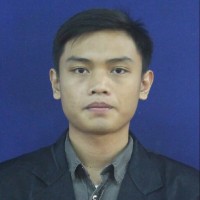}}]{AMAR FADILLAH} is a graduate of Bandung State Polytechnic, where he earned his Bachelor's degree in Industrial Automation Engineering in 2020. He further pursued his academic interests at National Taipei University of Technology, completing a Master's degree in International Program of Electrical Engineering and Computer Science in 2023. His research primarily focuses on Computer Vision, Autonomous Vehicles and Large Language Model, aiming to develop innovative solutions that enhance technology in these areas.
\end{IEEEbiography}

\begin{IEEEbiography}[{\includegraphics[width=1in,height=1.25in,clip,keepaspectratio]{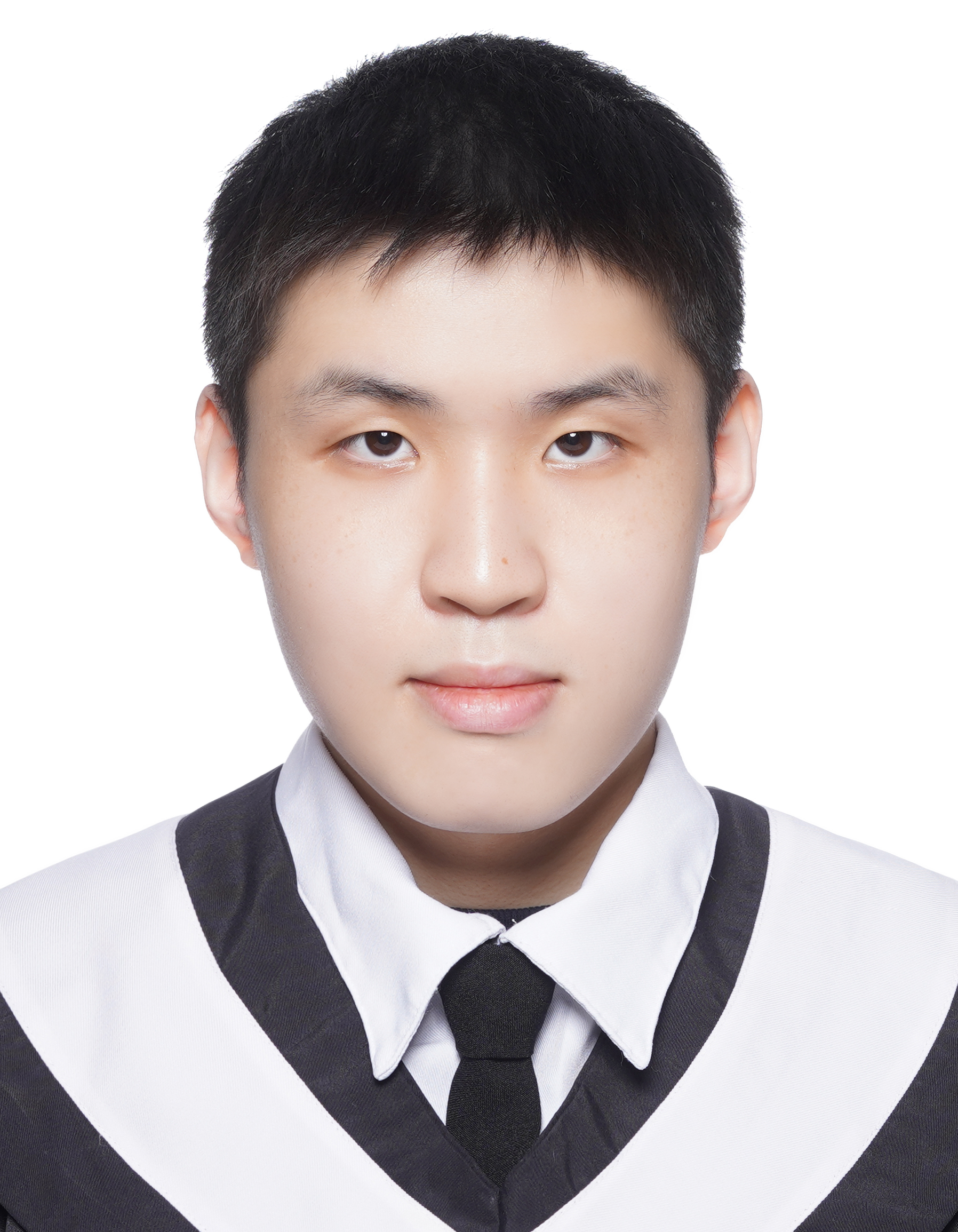}}]{Ching-Lin Lee} earned his Bachelor's and master's degree from National Taipei University of Science and Technology in 2022 and completed his Master's degree in 2024. His research primarily focuses on computer vision and Deep Learning.\end{IEEEbiography}

\begin{IEEEbiography}[{\includegraphics[width=1in,height=1.25in,clip,keepaspectratio]{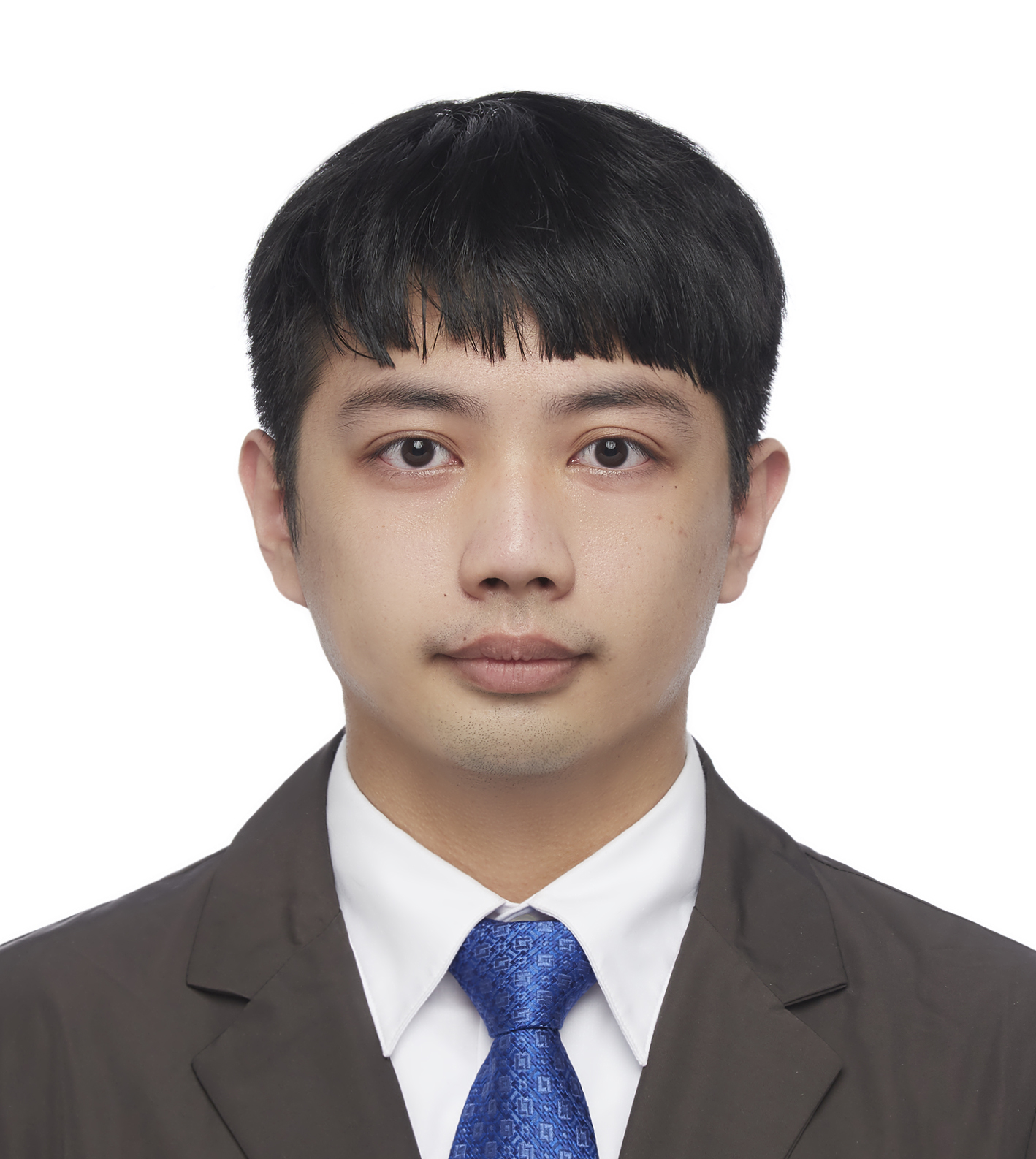}}]{Zhi-Xuan Wang} earned his Bachelor's degree in Applied Statistics from National Taichung University of Science and Technology in 2022 and completed his Master's degree in Artificial Intelligence at National Taipei University of Technology in 2024. His research primarily focuses on computer vision and autonomous vehicles.\end{IEEEbiography}

\begin{IEEEbiography}[{\includegraphics[width=1in,height=1.25in,clip,keepaspectratio]{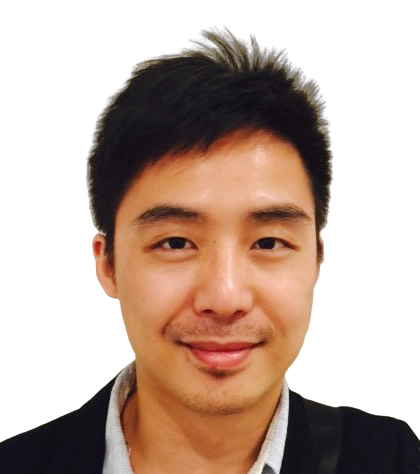}}]{Kuan-Ting Lai} received his B.Eng. degree in Electrical Engineering (2003), M.Sc. degree in Computer Science (2005), and Ph.D. degree (2015) from National Taiwan University. He became a Member (M) of IEEE in 2019. 

He was a visiting scholar in the DVMM lab at Columbia University from 2012 to 2013, and collaborated with IBM Thomas J. Watson Research Center to develop a large-scale video event detection system. He was a postdoctoral researcher with Academia Sinica, Taiwan, and led a group to win the ACM Multimedia Best Open-Source Software Award in 2018. 

He is currently an Associate Professor with the Department of Electronic Engineering, National Taipei University of Technology, Taiwan. His research interests include computer vision, deep learning, UAVs, and the Internet of Things.\end{IEEEbiography}

\EOD


\begin{thebibliography}{00}
\bibitem{kapal}Yu, Z., Du, J., Wang, J. \& Lin, Z. Prescribed-Performance Green Dynamic Positioning for Fully Actuated Vessels Under Input Magnitude and Rate Saturations. {\em IEEE Transactions On Automation Science And Engineering}. \textbf{22} pp. 14940-14952 (2025)
\bibitem{goal_add}Tran, H., Le, V. \& Tran, T. Goal-Driven Long-Term Trajectory Prediction. {\em Proceedings Of The IEEE/CVF Winter Conference On Applications Of Computer Vision (WACV)}. pp. 796-805 (2021,1)
\bibitem{rasouli2017JAAD}Rasouli, A., Kotseruba, I. \& Tsotsos, J. Are they going to cross? A benchmark dataset and baseline for pedestrian crosswalk behavior. {\em Proceedings Of The IEEE/CVF International Conference On Computer Vision (ICCV)}. pp. 206-213 (2017)
\bibitem{Rasouli2019PIE}Rasouli, A., Kotseruba, I., Kunic, T. \& Tsotsos, J. PIE: A Large-Scale Dataset and Models for Pedestrian Intention Estimation and Trajectory Prediction. {\em Proceedings Of The IEEE/CVF International Conference On Computer Vision (ICCV)}. (2019)
\bibitem{mangalam2021goals}Mangalam, K., An, Y., Girase, H. \& Malik, J. From goals, waypoints \& paths to long term human trajectory forecasting. {\em Proceedings Of The IEEE/CVF International Conference On Computer Vision (ICCV)}. pp. 15233-15242 (2021)
\bibitem{yao2021bitrap}Yao, Y., Atkins, E., Johnson-Roberson, M., Vasudevan, R. \& Du, X. Bitrap: Bi-directional pedestrian trajectory prediction with multi-modal goal estimation. {\em IEEE Robotics And Automation Letters}. \textbf{6}, 1463-1470 (2021)
\bibitem{wang2022stepwise}Wang, C., Wang, Y., Xu, M. \& Crandall, D. Stepwise goal-driven networks for trajectory prediction. {\em IEEE Robotics And Automation Letters}. \textbf{7}, 2716-2723 (2022)
\bibitem{pellegrini2010improving}Pellegrini, S., Ess, A. \& Van Gool, L. Improving data association by joint modeling of pedestrian trajectories and groupings. {\em Proceedings Of The European Conference On Computer Vision (ECCV)}. pp. 452-465 (2010)
\bibitem{lerner2007crowds}Lerner, A., Chrysanthou, Y. \& Lischinski, D. Crowds by example. {\em Proceedings Of The Computer Graphics Forum (CGF)}. pp. 655-664 (2007)
\bibitem{Robicquet2016learning}Robicquet, A., Sadeghian, A., Alahi, A. \& Savarese, S. Learning Social Etiquette: Human Trajectory Understanding In Crowded Scenes. {\em Proceedings Of The European Conference On Computer Vision (ECCV)}. pp. 549-565 (2016)
\bibitem{anderson2019stochastic}Anderson, C., Du, X., Vasudevan, R. \& Johnson-Roberson, M. Stochastic Sampling Simulation for Pedestrian Trajectory Prediction. {\em Proceedings Of The IEEE/RSJ International Conference On Intelligent Robots And Systems (IROS)}. pp. 4236-4243 (2019)
\bibitem{yuning2020multipath}Chai, Y., Sapp, B., Bansal, M. \& Anguelov, D. MultiPath: Multiple Probabilistic Anchor Trajectory Hypotheses for Behavior Prediction. {\em Proceedings Of The Conference On Robot Learning (CoRL)}. \textbf{100} pp. 86-99 (2020), https://proceedings.mlr.press/v100/chai20a.html
\bibitem{ivanovic2019trajectron}Ivanovic, B. \& Pavone, M. The trajectron: Probabilistic multi-agent trajectory modeling with dynamic spatiotemporal graphs. {\em Proceedings Of The IEEE/CVF International Conference On Computer Vision (ICCV)}. pp. 2375-2384 (2019)
\bibitem{salzmann2020trajectron++}Salzmann, T., Ivanovic, B., Chakravarty, P. \& Pavone, M. Trajectron++: Dynamically-feasible trajectory forecasting with heterogeneous data. {\em Proceedings Of The European Conference On Computer Vision (ECCV)}. pp. 683-700 (2020)
\bibitem{liu2022convnet}Liu, Z., Mao, H., Wu, C., Feichtenhofer, C., Darrell, T. \& Xie, S. A convnet for the 2020s. {\em Proceedings Of The IEEE/CVF Conference On Computer Vision And Pattern Recognition (CVPR)}. pp. 11976-11986 (2022)
\bibitem{wong2022view}Wong, C., Xia, B., Hong, Z., Peng, Q., Yuan, W., Cao, Q., Yang, Y. \& You, X. View vertically: A hierarchical network for trajectory prediction via fourier spectrums. {\em Proceedings Of The European Conference On Computer Vision (ECCV)}. pp. 682-700 (2022)
\bibitem{yue2022human}Yue, J., Manocha, D. \& Wang, H. Human trajectory prediction via neural social physics. {\em Proceedings Of The European Conference On Computer Vision (ECCV)}. pp. 376-394 (2022)
\bibitem{guo2022end}Guo, K., Liu, W. \& Pan, J. End-to-end trajectory distribution prediction based on occupancy grid maps. {\em Proceedings Of The IEEE/CVF Conference On Computer Vision And Pattern Recognition (CVPR)}. pp. 2242-2251 (2022)
\bibitem{Alahi2016Social}Alahi, A., Goel, K., Ramanathan, V., Robicquet, A., Fei-Fei, L. \& Savarese, S. Social LSTM: Human Trajectory Prediction in Crowded Spaces. {\em Proceedings Of The IEEE/CVF Conference On Computer Vision And Pattern Recognition (CVPR)}. pp. 961-971 (2016)
\bibitem{gupta2018social}Gupta, A., Johnson, J., Fei-Fei, L., Savarese, S. \& Alahi, A. Social gan: Socially acceptable trajectories with generative adversarial networks. {\em Proceedings Of The IEEE/CVF Conference On Computer Vision And Pattern Recognition (CVPR)}. pp. 2255-2264 (2018)
\bibitem{liang2020garden}Liang, J., Jiang, L., Murphy, K., Yu, T. \& Hauptmann, A. The garden of forking paths: Towards multi-future trajectory prediction. {\em Proceedings Of The IEEE/CVF Conference On Computer Vision And Pattern Recognition (CVPR)}. pp. 10508-10518 (2020)
\bibitem{liang2020simaug}Liang, J., Jiang, L. \& Hauptmann, A. SimAug: Learning Robust Representations from Simulation for Trajectory Prediction. {\em In Proceedings Of The European Confer- Ence On Computer Vision (ECCV)}. pp. 275-292 (2020)
\bibitem{levine2016end}Finn, C., Tan, X., Duan, Y., Darrell, T., Levine, S. \& Abbeel, P. Deep Spatial Autoencoders for Visuomotor Learning. {\em IEEE International Conference On Robotics And Automation (ICRA)}. pp. 512-519 (2016)
\bibitem{ronneberger2015u}Ronneberger, O., Fischer, P. \& Brox, T. U-net: Convolutional networks for biomedical image segmentation. {\em Proceedings Of The Medical Image Computing And Computer Assisted Intervention (MICCAI)}. pp. 234-241 (2015)
\bibitem{ioffe2017batch}Ioffe, S. Batch renormalization: Towards reducing minibatch dependence in batch-normalized models. {\em Advances In Neural Information Processing Systems}. \textbf{30} (2017)
\bibitem{wu2018group}Wu, Y. \& He, K. Group normalization. {\em Proceedings Of The European Conference On Computer Vision (ECCV)}. pp. 3-19 (2018)
\bibitem{ba2016layer}Ba, J., Kiros, J. \& Hinton, G. Layer normalization. {\em ArXiv Preprint ArXiv:1607.06450}. (2016)
\bibitem{chen2017deeplab}Chen, L., Papandreou, G., Kokkinos, I., Murphy, K. \& Yuille, A. Deeplab: Semantic image segmentation with deep convolutional nets, atrous convolution, and fully connected crfs. {\em Proceedings Of The IEEE Transactions On Pattern Analysis And Machine Intelligence}. \textbf{40}, 834-848 (2017)
\bibitem{he2016deep}He, K., Zhang, X., Ren, S. \& Sun, J. Deep residual learning for image recognition. {\em Proceedings Of The IEEE/CVF Conference On Computer Vision And Pattern Recognition (CVPR)}. pp. 770-778 (2016)
\bibitem{oktay2018attention}Oktay, O., Schlemper, J., Folgoc, L., Lee, M., Heinrich, M., Misawa, K., Mori, K., McDonagh, S., Hammerla, N., Kainz, B., Glocker, B. \& Rueckert, D. Attention U-Net: Learning Where to Look for the Pancreas. {\em Medical Imaging With Deep Learning}. (2018)
\bibitem{he2016identity}He, K., Zhang, X., Ren, S. \& Sun, J. Identity mappings in deep residual networks. {\em Proceedings Of The European Conference On Computer Vision (ECCV)}. pp. 630-645 (2016)
\bibitem{liu2023dropout}Liu, Z., Xu, Z., Jin, J., Shen, Z. \& Darrell, T. Dropout Reduces Underfitting. {\em Proceedings Of The 40th International Conference On Machine Learning (ICML)}. (2023)
\bibitem{girshick2015fast}Girshick, R. Fast r-cnn. {\em Proceedings Of The IEEE/CVF International Conference On Computer Vision (ICCV)}. pp. 1440-1448 (2015)
\bibitem{caesar2020nuscenes}Caesar, H., Bankiti, V., Lang, A., Vora, S., Liong, V., Xu, Q., Krishnan, A., Pan, Y., Baldan, G. \& Beijbom, O. nuscenes: A multimodal dataset for autonomous driving. {\em Proceedings Of The IEEE/CVF Conference On Computer Vision And Pattern Recognition (CVPR)}. pp. 11621-11631 (2020)
\bibitem{chang2019argoverse}Chang, M., Lambert, J., Sangkloy, P., Singh, J., Bak, S., Hartnett, A., Wang, D., Carr, P., Lucey, S., Ramanan, D. \& Others Argoverse: 3d tracking and forecasting with rich maps. {\em Proceedings Of The IEEE/CVF Conference On Computer Vision And Pattern Recognition (CVPR)}. pp. 8748-8757 (2019)
\bibitem{gu2023vip3d}Gu, J., Hu, C., Zhang, T., Chen, X., Wang, Y., Wang, Y. \& Zhao, H. ViP3D: End-to-end visual trajectory prediction via 3d agent queries. {\em Proceedings Of The IEEE/CVF Conference On Computer Vision And Pattern Recognition (CVPR)}. pp. 5496-5506 (2023)
\bibitem{wu2017modeling}Wu, H., Chen, Z., Sun, W., Zheng, B. \& Wang, W. Modeling Trajectories with Recurrent Neural Networks. {\em Proceedings Of The International Joint Conference On Artificial Intelligence (IJCAI)}. pp. 3083-3090 (2017)
\bibitem{park2018sequence}Park, S., Kim, B., Kang, C., Chung, C. \& Choi, J. Sequence-to-sequence prediction of vehicle trajectory via LSTM encoder-decoder architecture. {\em Proceedings Of The IEEE Intelligent Vehicles Symposium (IV)}. pp. 1672-1678 (2018)
\bibitem{kingma2014adam}Kingma, D. \& Ba, J. Adam: A Method for Stochastic Optimization. {\em International Conference On Learning Representations (ICLR)}. (2015)
\bibitem{xue2020poppl}Xue, H., Huynh, D. \& Reynolds, M. PoPPL: Pedestrian trajectory prediction by LSTM with automatic route class clustering. {\em IEEE Transactions On Neural Networks And Learning Systems}. \textbf{32}, 77-90 (2020)
\bibitem{tang2022evostgat}Tang, H., Wei, P., Li, J. \& Zheng, N. EvoSTGAT: Evolving spatiotemporal graph attention networks for pedestrian trajectory prediction. {\em Neurocomputing}. \textbf{491} pp. 333-342 (2022)
\bibitem{czech2022onboard}Czech, P., Braun, M., Kreßel, U. \& Yang, B. On-Board Pedestrian Trajectory Prediction Using Behavioral Features. {\em Proceedings Of The IEEE International Conference On Machine Learning And Applications (ICMLA)}. pp. 437-443 (2022)
\bibitem{abc2022}Halawa, M., Hellwich, O. \& Bideau, P. Action-Based Contrastive Learning for Trajectory Prediction. {\em Proceedings Of The European Conference On Computer Vision (ECCV)}. pp. 143-159 (2022)
\bibitem{9989439}Chen, W., Yang, Z., Xue, L., Duan, J., Sun, H. \& Zheng, N. Multimodal Pedestrian Trajectory Prediction Using Probabilistic Proposal Network. {\em IEEE Transactions On Circuits And Systems For Video Technology}. \textbf{33}, 2877-2891 (2023)
\bibitem{10685521}Lian, J., Wang, Z., Yang, D., Zheng, W., Li, L. \& Zhang, Y. Pedestrian Facial Attention Detection Using Deep Fusion and Multi-modal Fusion Classifier. {\em IEEE Transactions On Circuits And Systems For Video Technology}. pp. 1-1 (2024)
\bibitem{9153745}Chen, K., Song, X. \& Ren, X. Pedestrian Trajectory Prediction in Heterogeneous Traffic Using Pose Keypoints-Based Convolutional Encoder-Decoder Network. {\em IEEE Transactions On Circuits And Systems For Video Technology}. \textbf{31}, 1764-1775 (2021)
\bibitem{Liu2023}Liu, F., Duan, S. \& Juan, W. A pedestrian trajectory prediction method based on improved LSTM network. {\em IET Image Processing}. \textbf{18}, 379-387 (2023,10), http://dx.doi.org/10.1049/ipr2.12954
\bibitem{Zhou2023}Zhou, J., Bai, X., Fu, W., Ning, B. \& Li, R. Pedestrian intention estimation and trajectory prediction based on data and knowledge‐driven method. {\em IET Intelligent Transport Systems}. \textbf{18}, 315-331 (2023,11), http://dx.doi.org/10.1049/itr2.12453
\end{thebibliography}
\end{document}